\documentclass[journal,twocolumn]{IEEEtran}
\usepackage{graphicx,times,amsmath,amssymb,cite,subfigure,stfloats,booktabs, url,multirow,afterpage}
\usepackage[lined,ruled]{algorithm2e}
\allowdisplaybreaks

\usepackage{bbding}
\usepackage[normalem]{ulem}
\useunder{\uline}{\ul}{}

\makeatletter

\newcommand{\Rmnum}[1]{\expandafter\@slowromancap\romannumeral #1@}
\makeatother

\begin{document}
\title{Federated Motor Imagery Classification for Privacy-Preserving Brain-Computer Interfaces}

\author{Tianwang~Jia, Lubin~Meng, Siyang~Li, Jiajing~Liu and Dongrui~Wu
\thanks{T. Jia, L. Meng, S. Li and D. Wu are with the Key Laboratory of the Ministry of Education for Image Processing and Intelligent Control, School of Artificial Intelligence and Automation, Huazhong University of Science and Technology, Wuhan 430074, China. They are also with Shenzhen Huazhong University of Science and Technology Research Institute, Shenzhen 518063, China. J. Liu is with the School of Civil and Hydraulic Engineering, Huazhong University of Science and Technology, Wuhan 430074, China. Email: \{twjia, lubinmeng, syoungli, liu\_jiajing, drwu\}@hust.edu.cn.}}

\maketitle

\begin{abstract}
Training an accurate classifier for EEG-based brain-computer interface (BCI) requires EEG data from a large number of users, whereas protecting their data privacy is a critical consideration. Federated learning (FL) is a promising solution to this challenge. This paper proposes Federated classification with local Batch-specific batch normalization and Sharpness-aware minimization (FedBS) for privacy protection in EEG-based motor imagery (MI) classification. FedBS utilizes local batch-specific batch normalization to reduce data discrepancies among different clients, and sharpness-aware minimization optimizer in local training to improve model generalization.
Experiments on three public MI datasets using three popular deep learning models demonstrated that FedBS outperformed six state-of-the-art FL approaches. Remarkably, it also outperformed centralized training, which does not consider privacy protection at all. In summary, FedBS protects user EEG data privacy, enabling multiple BCI users to participate in large-scale machine learning model training, which in turn improves the BCI decoding accuracy.
\end{abstract}


\IEEEpeerreviewmaketitle

\section{Introduction}

A brain-computer interface (BCI)~\cite{Wolpaw2002} enables direct communication between the human brain and external devices. Electroencephalography (EEG) stands as one of the most commonly utilized input signals in non-invasive BCIs. This paper focuses on EEG-based motor imagery (MI) classification~\cite{Pfurtschelle2001}, a classical BCI paradigm. MI induces changes in EEG energy distribution by mentally imagining the movement of a body part without actually performing it.

EEG data from many subjects are usually needed to train an MI classifier with good generalization~\cite{Wu2022}. However, recent studies have found that EEG-based BCIs are subject to privacy threats~\cite{Xia2022}, e.g., EEG data could leak users' private information including personal preference, health status, mental state, and so on. Due to legal regulations and user concerns, privacy-preserving machine learning for BCIs becomes a necessity.

\begin{figure}[htbp]     \centering
    \includegraphics[width=0.95\linewidth]{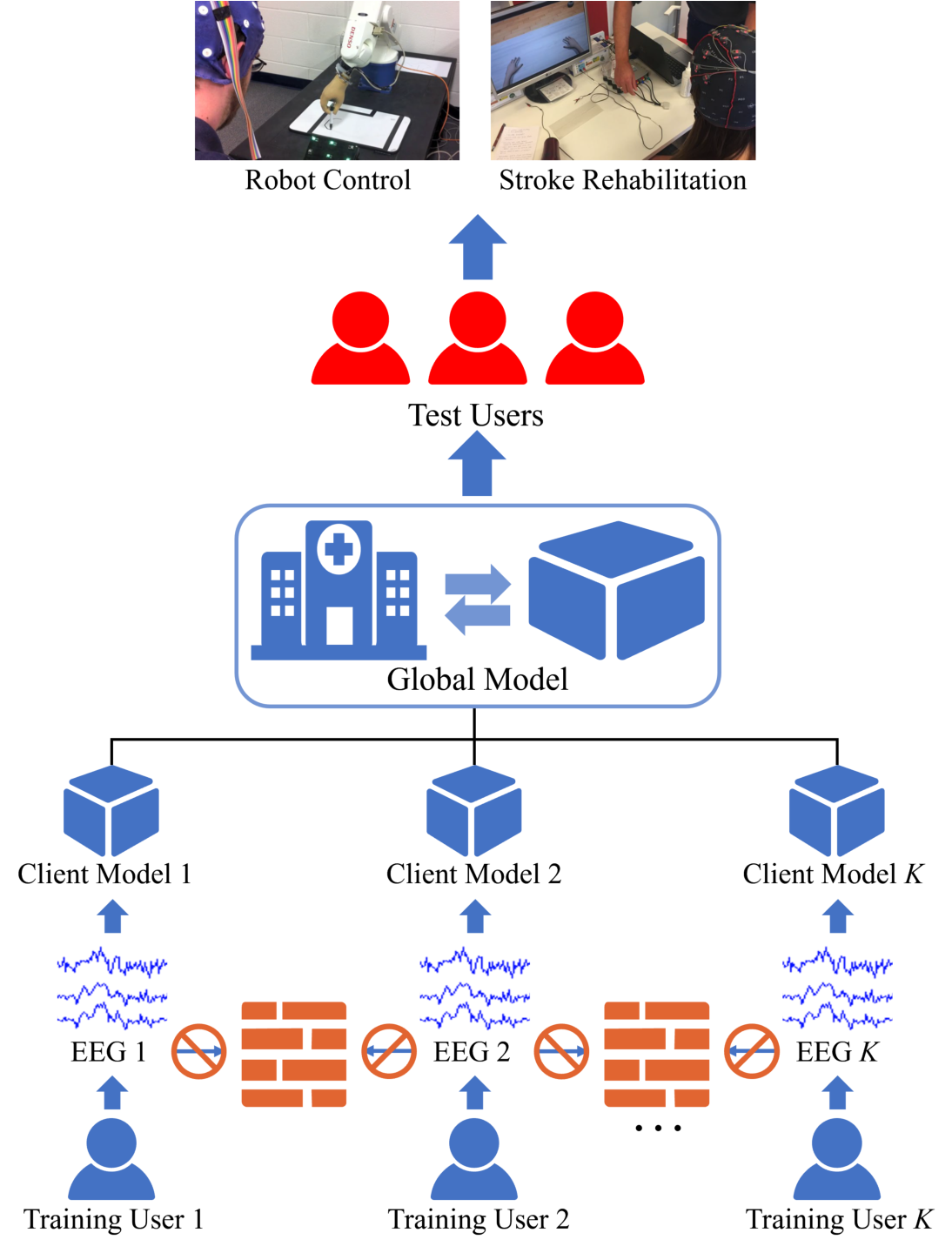}
    \caption{FedBS for privacy-preserving BCIs.}     \label{fig:app}
\end{figure}

Federated learning (FL)~\cite{Yang2019,Li2023} is a promising solution. For privacy-preserving BCIs, FL works as follows [illustrated using our proposed Federated classification with local Batch-specific batch normalization and Sharpness-aware minimization (FedBS) in Fig.~\ref{fig:app}]: a central server, which has no access to the local clients' private EEG data, maintains a global model and sends it to the local clients for updating; each client updates the global model parameters based on its own EEG data and sends them to the server for aggregation. In this way, a global model can be trained without sharing EEG data between the server and the clients, or among the clients. FL protects user privacy by preventing other devices from accessing raw data stored on the local client, thus avoiding the privacy risks of centralized datasets.

FedAvg~\cite{Mcmahan2017} is one of the most popular FL approaches. To reduce the communication overhead, FedAvg performs multiple stochastic gradient descent updates on some chosen clients in each communication round and then aggregates the models on the server until convergence. FedBN~\cite{Li2021} uses local batch normalization (BN) ~\cite{Ioffe2015} and excludes the BN layer parameters of all client models in communication, to alleviate client shifts; however, FedBN does not yield a complete classifier at the server, and is unable to adapt to previously unseen client data distributions.

FedBS improves the BN layer in FedAvg, ensuring each client's BN layers remain private during training, while the server has a complete model during the testing phase. It calculates batch-specific statistics for BN to mitigate feature shift across different clients. Furthermore, FedBS introduces the sharpness-aware minimization (SAM) optimizer~\cite{Foret2021} into the local training of the clients, encouraging their models to converge to flatter minima and hence enhancing the generalization. Experiments using three popular deep learning models on three MI datasets demonstrated FedBS's superior performance over six state-of-the-art FL approaches, even surpassing centralized training which does not consider privacy protection.

The remainder of this paper is organized as follows. Section~\ref{sect:related} introduces related works on privacy-preserving learning and FL. Section~\ref{sect:fedbs} proposes FedBS. Section~\ref{sect:experiments} shows the experimental results on three MI datasets. Section~\ref{sect:disscussion} presents some discussions and points out future research directions. Finally, Section~\ref{sect:conclusion} draws conclusions.

\section{Background Information on Privacy-preserving Machine Learning} \label{sect:related}

This section introduces background knowledge and related works on privacy-preserving machine learning and FL, which will be used in the next section.

\subsection{Privacy Protection}

Privacy protection is especially important in BCI applications. Developing a commercial BCI system may require collaboration among multiple organizations, e.g., hospitals, universities, and/or companies. If raw EEG data are transferred directly during this process, then private information, e.g., physical/emotional states, may be leaked~\cite{Xia2022}. With increasing privacy protection requirements from both the governments (e.g., the European General Data Protection Regulation, the American Data Privacy and Protection Act, and the Personal Information Protection Law of China) and the end-users, privacy protection becomes necessary.

There are four popular privacy protection strategies~\cite{Al2019,Xu2021}: cryptography, perturbations, source-free domain adaptation, and FL. Cryptography utilizes encryption techniques like homomorphic encryption~\cite{Kuri2017} and secure multi-party computation~\cite{Ahlswede1993} to protect data privacy. Perturbation techniques such as differential privacy~\cite{Agarwal2019} add noise to or alter the original data while maintaining their utility for downstream tasks. The other two strategies are introduced in more detail next.

\subsection{Source-free Domain Adaptation}

Source-free domain adaptation~\cite{Liang2021}, a subcategory of transfer learning~\cite{drwuTLBCI2022}, considers adapting to a target domain (test subject) that has data distribution shift from the source domain (training subject). In contrast to traditional transfer learning, source-free transfer learning performs adaptation without accessing the source domain data, usually in the form of transferring a trained source model, to ensure source data privacy protection.

For EEG-based BCIs, Xia \emph{el al.}~\cite{Xia2022a} proposed augmentation-based source-free domain adaptation for cross-subject MI classification. Zhang \emph{et al.} proposed lightweight source-free transfer~\cite{Zhang2022LSFT}, and further studied multi-source decentralized transfer~\cite{Zhang2022MSDT}, for privacy-preserving BCIs.

\subsection{Federated Learning}

FL aims to build a global model from private data located at multiple sites, without access to the raw data.

FedAvg, a simple yet widely used FL approach, has difficulty in handling local data heterogeneity. To tackle this problem, FedProx~\cite{Li2020} introduces a proximal term into the local objective function of the clients to reduce the discrepancy between local and global models. SCAFFOLD~\cite{Karimireddy2020} employs control variables to correct the ``client drift" during local updates. Hsu~\emph{et al.}~\cite{Hsu2019} introduced the concept of momentum into server-side aggregation to mitigate distribution discrepancies. Some works have integrated optimization techniques with federated learning. Reddi~\emph{et al.}~\cite{Reddi2021} introduced federated versions of adaptive optimizers (Adagrad, Adam, and Yogi), which improve federated learning convergence speed and performance. Jin~\emph{et al.}~\cite{Jin2022} proposed FedDA, a momentum-decoupling adaptive optimization approach, ensuring convergence in federated learning.

Limited studies have been carried out on FL for BCIs, and their primary goal was not privacy protection. Ju \emph{et al.}~\cite{Ju2020} introduced a federated transfer learning framework, which uses single-trial covariance matrices and domain adaptation techniques to extract shared discriminative information from EEG data of multiple subjects; however, their approach may compromise user data privacy and does not work with Euclidean space features and models. Liu \emph{et al.}~\cite{Liu2023} split the classifier into a local module and a global module to merge knowledge from different EEG datasets; however, they mainly focused on personalized FL, with limited attention to user data privacy protection.

In summary, there lacks a generic FL approach for BCIs, which is applicable to Euclidean space features, achieves user data privacy protection, and boosts the decoding performance simultaneously. Compatibility with Euclidean space features means that the approach can work with diverse network architectures commonly used in EEG data analysis, and be seamlessly integrated with various data preprocessing, data augmentation, and machine learning algorithms.

\section{FedBS} \label{sect:fedbs}

This section introduces the details of our proposed FedBS approach.

\subsection{Definitions and Notations}

Table~\ref{tab:notation} summarizes the main notations used in this paper.

\begin{table}[htbp]    \centering    \setlength{\tabcolsep}{3.0mm}
    \caption{Main notations.}    \label{tab:notation}
    \begin{tabular}{l|l}        \toprule
        Notation & Description                            \\        \midrule
        $\mathcal{D}_{c,k}$    & Domain of the $k$-th client                       \\
        $\mathcal{D}_s$    & Domain of server                       \\
        $X_i\in \mathbb{R}^{C\times T}$      & The $i$-th EEG trial                           \\
        $y_i\in \{1,\dots,N_c\}$      & Label of the $i$-th EEG trial                     \\
        $\mathbf{w}$      & Classifier parameters                       \\
        $K$      & Number of clients                      \\
        $C$      & Number of EEG channels                 \\
        $T$      & Number of time sampling points         \\
        $n$      & Number of EEG trials                   \\
        $N_c$    & Number of classes                      \\
        $N_t$    & Number of maximum communication rounds \\    \bottomrule
    \end{tabular}
\end{table}\textbf{}

Assume there are $K$ subjects as clients, and the $k$-th client has $n_k$ EEG trials $\mathcal{D}_{c,k}=\{(X_i,y_i)\}_{i=1}^{n_k}$ and a local classifier with parameters $\mathbf{w}_t^k$, where $X_i \in \mathbb{R}^{C\times T}$ ($C$ is the number of EEG channels, and $T$ the number of sampling points), $y_i\in \{1,\dots,N_c\}$ ($N_c$ is the number of classes) and $t\in \{1,\dots, N_t\}$ ($N_t$ is the number of maximum communication rounds) respectively represent the $i$-th trial, its corresponding label and the communication rounds. Assume also there is a server with $n_s$ EEG trials $\mathcal{D}_s=\{(X_i,y_i)\}_{i=1}^{n_s}$ from an unknown subject for test. The goal is to train a classifier without exposing the raw client data and subsequently evaluate its performance on the test subjects at the server.

\subsection{Overview of FedBS}

\begin{figure*}[htbp]    \centering
    \includegraphics[width=0.9\textwidth]{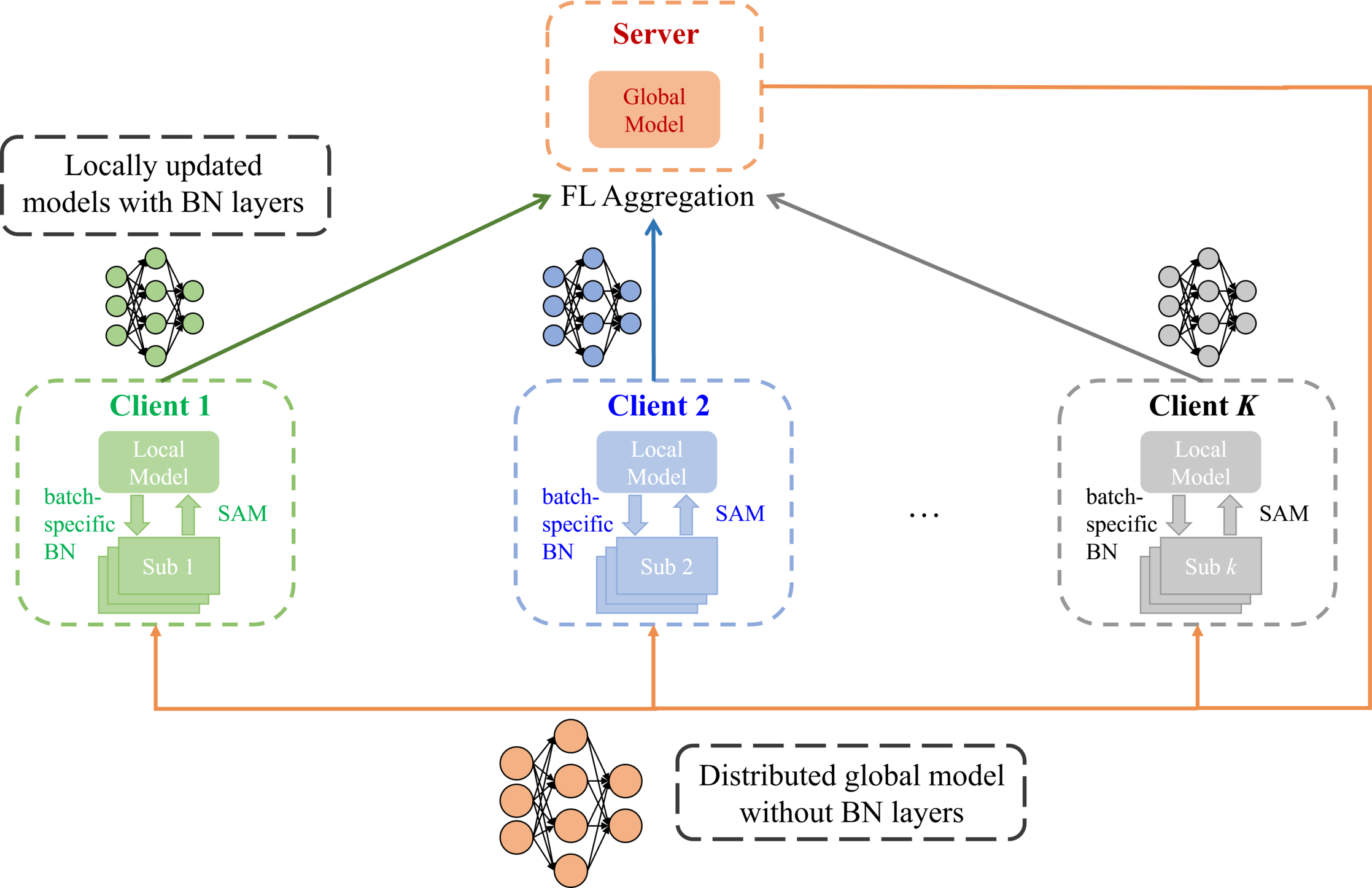}
    \caption{Overview of FedBS. Each client represents an individual subject. `Batch-specific BN' means the BN layer statistics are computed independently for each batch.}    \label{fig:framework}
\end{figure*}

FedBS consists of a server and several clients, as illustrated in Fig.~\ref{fig:framework}. Algorithm~\ref{alg:fedbs} shows the pseudo-code. The Python code is available at \url{https://github.com/TianwangJia/FedBS}.

Each client holds data from an individual subject and trains a local classifier on it. The server handles model distribution and aggregation, with BN layers added only during aggregation. We utilize local batch-specific BN to reduce feature shift across different clients, as described in Section~\ref{subsec:bn}. Furthermore, we introduce an SAM optimizer for client model training, as outlined in Section~\ref{subsec:sam}.

\begin{algorithm}[htbp]
    \KwIn{$K$, number of clients\;
        \hspace*{9mm} $D_k$, labeled dataset of the $k$-th client\;
        \hspace*{9mm} $N_t$, maximum communication rounds\;
        \hspace*{9mm} $\eta$, learning rate\;
        \hspace*{9mm} $P$, client selection weight\;
        \hspace*{9mm} $B$, batch size\;
        \hspace*{9mm} $E$, number of local optimization epochs on each client\;
        \hspace*{9mm} $\rho$, gradient ascent step size for SAM in (\ref{eq:sam_optim}).}
    \KwOut{$\mathbf{w}$, model parameters}
    ~\\
    \textbf{Server executes:}\\
    Initialize $\mathbf{w}_0$\;
    \For{$t=1,\dots,N_t$}{
        $m\leftarrow \max(P\cdot K,1)$\;
        $S_t\leftarrow \text{Set of $m$ randomly selected clients}$\;
        \For{client $k \in S_t$}{
                \tcp{Server does not distribute BN layer parameters;}
                $\mathbf{w}_t\leftarrow \text{Model parameters excluding BN layers}$\;
                $\mathbf{w}^k_{t+1}\leftarrow \text{ClientUpdate}(k,\mathbf{w}_t)$\;
        }
        $\mathbf{w}_{t+1}\leftarrow \sum_{k\in S_t}\frac{n_k}{\sum_{k\in S_t}n_k} \mathbf{w}^k_{t+1}$\;
    }
    ~\\
    \textbf{ClientUpdate$(k,\mathbf{w})$:}\\
    $\mathbb{B}\leftarrow \text{Split $D_k$ into batches of size $B$}$\;
    \For{$\text{epoch}=1,\dots,E$}{
        \For{$\text{batch }\mathcal{B} \in \mathbb{B}$}{
            Calculate $\mu_{\mathcal{B}}$ on $\mathcal{B}$ by (\ref{eq:bn_mu});\

            Calculate $\sigma_{\mathcal{B}}$ on $\mathcal{B}$ by (\ref{eq:bn_sigma});\

            Update $\mathbf{w}$ with $(\mu_{\mathcal{B}},\sigma_{\mathcal{B}})$ by (\ref{eq:sam_optim});\
        }
    }
    \caption{FedBS} \label{alg:fedbs}
\end{algorithm}

\subsection{Local Batch-Specific BN} \label{subsec:bn}

Inspired by FedBN~\cite{Li2021}, we localize the BN layer parameters and improve the BN's FL training to accommodate generalization to test scenarios:
\begin{itemize}
    \item Clients upload all parameters, including those of the BN layer;
    \item Server aggregates all model parameters, but distributes model parameters without those of the BN layer.
\end{itemize}

This approach ensures localized BN parameters on each client for better adaptation to its specific data distribution, while also providing the server with a complete model structure and model parameters for on-demand tests.

To better accommodate the EEG data distribution discrepancies from different subjects, FedBS further calculates batch-specific statistics of BN.

As pointed out in~\cite{Ioffe2015}, the BN layer has four sets of parameters: $\mu$, $\sigma$, $\gamma$, and $\beta$. $\mu$ and $\sigma$ are respectively the mean and variance, non-trainable parameters calculated during the forward propagation process. $\gamma$ and $\beta$ are affine transformation parameters, trainable during the backward propagation process.

In prior works, $\mu$ and $\sigma$ are computed over the entire training process and remain fixed during tests, assuming training and test data have similar distributions. However, in our setting, different clients generally have different data distributions; hence, global statistics in training may not adapt well to feature shift on different clients during training, and also new subjects during tests.

As a result, FedBS calculates batch-specific statistics. Specifically, for a batch of training or test data $\mathcal{B}=\{(X_i,y_i)\}_{i=1}^B$, it calculates
\begin{eqnarray}
\mu_{\mathcal{B}}&= &\frac{1}{B}\sum_{i=1}^{B}X_i,\label{eq:bn_mu}\\
\sigma^2_{\mathcal{B}}&= & \frac{1}{B}\sum_{i=1}^{B}(X_i-\mu_{\mathcal{B}})^2.\label{eq:bn_sigma}
\end{eqnarray}
Batch-specific statistics computation minimizes feature shift among clients during training, and facilitates rapid adaptation to new data distributions during tests.

Note again that our proposed FedBS yields a complete classifier at the server, and is able to adapt to previously unseen client data distributions, whereas FedBN~\cite{Li2021} is primarily for personalized FL scenarios.

\subsection{SAM Optimizer} \label{subsec:sam}

To boost model generalization, inspired by~\cite{Caldarola2022,Qu2022}, we employ the SAM optimizer~\cite{Foret2021} in client model training.

As pointed out in~\cite{Caldarola2022}, client shift leads each local model to focus on its own biased data, deviating from the global optimum. SAM helps each local model converge to a flatter optimum in the loss landscape. When aggregated, the global model could be closer to the global optimum and have better generalization.

Specifically, SAM minimizes the following loss function:
\begin{align}
    \min_\mathbf{w}\ \  \mathcal{L}^{SAM}(\mathbf{w})+\frac{\lambda}{2} \Vert \mathbf{w}\Vert^2_2, \label{eq:sam_obj}
\end{align}
where $\mathcal{L}^{SAM}(\mathbf{w})\triangleq\max_{\Vert \boldsymbol{\epsilon} \Vert_p\leq \rho} \mathcal{L}(\mathbf{w}+\boldsymbol{\epsilon})$ is the SAM loss, in which $\mathcal{L}(\mathbf{w})$ is the cross-entropy loss of the original parameters, $\mathcal{L}(\mathbf{w}+\boldsymbol{\epsilon})$ is the cross-entropy loss when parameters are adjusted by $\boldsymbol{\epsilon}$, $\rho \geq 0$ is a hyperparameter that governs the range of $\boldsymbol{\epsilon}$, and $p\in [1,\infty]$ ($p=2$ is often used~\cite{Foret2021}). $\frac{\lambda}{2} \Vert \mathbf{w} \Vert_2^2$ is a regularization.

$\mathcal{L}^{SAM}(\mathbf{w})$ above can be re-expressed as:
\begin{align}
\mathcal{L}^{SAM}(\mathbf{w})=\mathcal{L}(\mathbf{w})+\left[\max_{\Vert \epsilon \Vert_p\leq \rho} \mathcal{L}(\boldsymbol{\omega}+\boldsymbol{\epsilon})-\mathcal{L}(\mathbf{w})\right]. \label{eq:Lsam}
\end{align}
The first term is the cross-entropy loss, whereas the latter is the sharpness of the cross-entropy loss. Thus, SAM simultaneously minimizes both the cross-entropy loss (classification loss) and its sharpness, leading to better generalization.

To solve (\ref{eq:Lsam}), we first find $\boldsymbol{\epsilon}^*(\mathbf{w})$ that maximizes $\mathcal{L}(\mathbf{w}+\boldsymbol{\epsilon})$:
\begin{align}
    \begin{split}
        \boldsymbol{\epsilon}^*(\mathbf{w})&\triangleq \mathop{\arg\max}_{\Vert \boldsymbol{\epsilon} \Vert_p\leq \rho}\mathcal{L}(\mathbf{w}+\boldsymbol{\epsilon})  \\
        &\approx \mathop{\arg\max}_{\Vert \boldsymbol{\epsilon} \Vert_p\leq \rho} \left(\mathcal{L}(\mathbf{w})+\boldsymbol{\epsilon}^\top\nabla_\mathbf{w}\mathcal{L}(\mathbf{w})\right ) \\
        &=\mathop{\arg\max}_{\Vert \boldsymbol{\epsilon} \Vert_p\leq \rho}\boldsymbol{\epsilon}^\top\nabla_\mathbf{w}\mathcal{L}(\mathbf{w}). \label{eq:sam_eps}
    \end{split}
\end{align}
When the $L_2$ norm is used, (\ref{eq:sam_eps}) yields:
\begin{align}
    \boldsymbol{\epsilon}^*(\mathbf{w})=\rho \frac{\nabla_\mathbf{w}\mathcal{L}(\mathbf{w})}{\Vert\nabla_\mathbf{w}\mathcal{L}(\mathbf{w})\Vert_2}.
\end{align}

Substituting $\boldsymbol{\epsilon}^*(\mathbf{w})$ into (\ref{eq:Lsam}) and  neglecting the higher-order terms, we have:
\begin{align}
\nabla_\mathbf{w}\mathcal{L}^{SAM}(\mathbf{w})
\approx\nabla_\mathbf{w}\mathcal{L}(\mathbf{w}+\boldsymbol{\epsilon}^*(\mathbf{w}))\approx \nabla_\mathbf{w}\mathcal{L}(\mathbf{w})\vert_{\mathbf{w}+\boldsymbol{\epsilon}^*\left(\mathbf{w}\right)}.
\end{align}

Let $\mathbf{w}_t$ be the classifier parameters after $t$ steps of model update. Then, SAM involves two steps:
\begin{align}
    \left\{
    \begin{aligned}
        &\boldsymbol{\epsilon}^*_t=\rho \frac{\nabla_{\mathbf{w}_t}\mathcal{L}(\mathbf{w}_t)}{\Vert \nabla_{\mathbf{w}_t}\mathcal{L}(\mathbf{w}_t)\Vert_2}\\
        &\mathbf{w}_{t+1} = \mathbf{w}_t-\eta \left(\nabla_{\mathbf{w}_t}\mathcal{L}(\mathbf{w}_t+\boldsymbol{\epsilon}^*_t)+\lambda \mathbf{w}_t \right)
    \end{aligned}
    \right..
    \label{eq:sam_optim}
\end{align}
So, SAM performs first gradient ascent to find model parameters that maximize the loss function within a specified range, and then gradient descent on the original model parameters.

\section{Experiments and Results} \label{sect:experiments}

We performed MI classification experiments on three EEG datasets to validate the effectiveness of FedBS.

\subsection{Datasets and Preprocessing}

Three EEG-based MI datasets, summarized in Table~\ref{tab:dataset}, were used in our experiments. Theoretical chance levels for the three datasets are 25\%, 50\%, and 50\%, respectively~\cite{Combrisson2015}.

\begin{table}[htbp]    \centering    \setlength{\tabcolsep}{3.0mm}
    \caption{Summary of the three MI datasets.}    \label{tab:dataset}
    \begin{tabular}{c|cccc}        \toprule
        Dataset   & \# Subjects & \# Classes & \# Channels & \# Trials \\
        \midrule
        MI1~\cite{Tangermann2012} & 9           & 4          & 22          & 576       \\
        MI2~\cite{Steyrl2016} & 14          & 2          & 15          & 160       \\
        MI3~\cite{Faller2012} & 12          & 2          & 13          & 400      \\
        \bottomrule
    \end{tabular}
\end{table}

The three datasets used similar data collection procedure. A subject sat in front of a computer screen. Each trial began with a fixation cross and a brief warning tone. Following that a visual cue (e.g., an arrow) appeared for several seconds, during which the subject performed specific a MI task. Then, there was a brief rest period. EEG signals were recorded throughout the entire experiment.

MI1 was the BNCI 2014-001 MI dataset~\cite{Tangermann2012}. It includes four classes (left hand, right hand, both feet, and tongue), each with 144 trials. 22-channel EEG signals from 9 healthy subjects were recorded at 250 Hz.

MI2 was the BNCI 2014-002 MI dataset~\cite{Steyrl2016}. It includes two classes (left hand versus right hand), each with 80 trials. 15-channel EEG signals from 14 healthy subjects were sampled at 512Hz.

MI3 was the BNCI 2015-001 MI dataset~\cite{Faller2012}. It has two classes (right hand versus both feet), each with 200 trials. 13-channel EEG signals from 12 healthy subjects were sampled at 250Hz.

All datasets were downloaded and 8-30Hz bandpass filtered using MOABB~\cite{Jayaram2018}. MI2 was also downsampled to 250Hz, to be consistent with the other two datasets. We extracted EEG trials between [0,4]s for MI1, and [0,5]s for MI2 and MI3, after each task stimulus.

\subsection{Baseline Algorithms}

We compared FedBS with a centralized training approach and six existing FL approaches:
\begin{enumerate}
    \item Centralized training (CT), which pools data from all subjects together for training, without any privacy protection.
    \item FedAvg~\cite{Mcmahan2017}, the most widely used FL algorithm. Each client performs multiple stochastic gradient descent update steps in a single communication round to balance the communication cost and the accuracy.
    \item FedProx~\cite{Li2020}, which introduces a proximal term into the client objective function to reduce the discrepancy between local and global models.
    \item SCAFFOLD~\cite{Karimireddy2020}, which employs control variables to correct the `client drift' during local updates.
    \item MOON~\cite{Li2021a}, which leverages the similarity between model representations in local training.
    \item FedFA~\cite{Zhou2023}, which performs federated feature augmentation.
    \item GA~\cite{Zhang2023}, which introduces a fairness objective measured by the variance of the generalization gaps among different source domains, and optimizes it through dynamic aggregation weight adjustments. Note that GA requires all clients to participate in every communication round.
\end{enumerate}

\subsection{Experiment Settings and Hyperparameters}

We used leave-one-subject-out cross-validation in performance evaluation. Six repeats with different random seeds were performed, and the average results are reported.

Specifically, in CT, one subject was designated as the test subject, and all others' data were combined for model training. In all FL approaches, the number of clients equaled the total number of subjects minus one. Each client represented one training subject, and the server used the remaining subject's data for test. During each communication round, half of the clients (rounded down) were randomly chosen, and each performed two local training epochs.

Three backbone classifiers, EEGNet~\cite{Lawhern2018}, DeepConvNet~\cite{Schirrmeister2017} and ShallowConvNet~\cite{Schirrmeister2017}, were used. The hyperparameters for EEGNet were: $F_1=8$, $D=2$, and $F_2=16$. The network architectures and other parameters of the three models aligned with their respective literature and can be further fine-tuned according to the reference~\cite{Cooney2020}. We performed Euclidean alignment (EA)~\cite{He2020} for each subject.

Stochastic gradient descent optimizer with weight decay 0.0001 and momentum 0.9 were used in all approaches. CT was trained for 200 epochs, whereas all others were trained for 200 global communication rounds. On MI1 and MI3, CT used a batch size of 64 and all others  used a batch size of 32; on MI2, since each subject had fewer samples, the batch sizes were 32 and 16, respectively. EEGNet and DeepConvNet used a learning rate of 0.005, whereas ShallowConvNet used 0.0001. The test batch size was fixed at 8.

FedBS introduces only one additional hyperparameter, which was set to $\rho=0.1$. Other hyperparameters for different FL approaches were fine-tuned within a small range based on their respective literature. $\mu = 1.0$ for FedProx, $\mu = 1.0$ and $\tau = 0.5$ for MOON, $p = 0.5$ for FedFa, and $d = 0.05$ for GA.

\subsection{Cross-Subject Classification Performance}

The cross-subject classification accuracies without EA are shown in Table~\ref{tab:MI1&2&3_withouEA}. The detailed results on the three MI datasets with EA are shown in Tables~\ref{tab:MI1_result2}-\ref{tab:MI3_EA}, respectively. The best performance in each column is marked in bold, and the second-best underlined. We can observe that:
\begin{enumerate}
	\item \emph{FedBS almost always achieved the best performance.} Remarkably, it almost always outperformed CT, which did not consider privacy protection at all. In other words, our proposed FedBS achieved simultaneously data privacy protection and classification accuracy improvement.
	\item \emph{EA boosted the performance of all approaches, including FedBS.} On average, when EA was used, FedBS outperformed CT by 1.97\%, and the second-best FL approaches by 3.08\%.
\end{enumerate}

To evaluate whether FedBS outperformed other approaches significantly, we first calculated the $p$-values of paired $t$-tests, and then adjusted them by Benjamini-Hochberg False Discovery Rate correction~\cite{Benjamini1995}. The results are shown in Table~\ref{tab:p-values}, where $p$-values smaller than $0.05$ are marked in bold. It is evident that the performance improvements of FedBS over others were almost always statistically significant. Further details of the paired $t$-tests are provided in Appendix.

\begin{table*}[htbp]	\centering \setlength{\tabcolsep}{5mm}
	\caption{Average cross-subject classification accuracies (\%) without EA. The best accuracy in each column is marked in bold, and the second best by an underline.}	\label{tab:MI1&2&3_withouEA}
	\begin{tabular}{c|ccc|ccc|ccc}
		\toprule
		\multirow{2}{*}{Approach} & \multicolumn{3}{c|}{EEGNet} & \multicolumn{3}{c|}{DeepConvnet} & \multicolumn{3}{c}{ShallowConvNet} \\ \cmidrule{2-10}
		& MI1     & MI2     & MI3     & MI1       & MI2       & MI3      & MI1        & MI2       & MI3       \\ \midrule
		CT          & {\ul44.42}   & {\ul65.05}   & {\ul63.99}   & {\ul44.85}     & {\ul69.54}     & {\ul63.98}    & 43.36      & {\ul63.05}     & {\ul63.33}     \\ \midrule
		FedAvg                  & 41.07   & 55.45   & 60.41   & 40.81     & 57.75     & 56.88    & {\ul43.81}      & 54.38     & 58.20     \\
		FedProx                 & 40.72   & 54.79   & 59.94   & 39.86     & 58.61     & 55.69    & 43.18      & 52.57     & 58.96     \\
		SCAFFOLD                & 30.69   & 53.25   & 59.76   & 28.55     & 51.15     & 55.08    & 35.94      & 53.53     & 59.31     \\
		MOON                    & 32.72   & 50.35   & 59.55   & 39.80     & 60.89     & 59.15    & 42.82      & 55.92     & 58.74     \\
		FedFA                   & 40.04   & 52.99   & 60.78   & 40.55     & 53.57     & 60.43    & 41.68      & 53.85     & 62.12     \\
		GA                      & 40.90   & 52.01   & 63.61   & 40.98     & 57.23     & 62.96    & 43.63      & 58.83     & 63.23     \\
		FedBS                   & \textbf{51.34}   & \textbf{73.80}   & \textbf{75.66}   & \textbf{51.20}     & \textbf{74.91}     & \textbf{74.66}    & \textbf{46.60}      & \textbf{66.79}     & \textbf{70.88}     \\	\bottomrule
	\end{tabular}
\end{table*}

\begin{table*}[htbp]    \centering \setlength{\tabcolsep}{3mm}
    \caption{Cross-subject classification accuracies (\%) with EA on MI1. The best accuracy in each column is marked in bold, and the second best by an underline.}
    \label{tab:MI1_result2}
    \begin{tabular}{c|c|ccccccccc|c}        \toprule
        \multirow{2}{*}{Model} & \multirow{2}{*}{Approach} & \multicolumn{9}{c|}{Subject}                                          & \multirow{2}{*}{Avg.} \\ \cmidrule{3-11}
        &                         & 1     & 2     & 3     & 4     & 5     & 6     & 7     & 8     & 9     &                          \\ \midrule
        \multirow{8}{*}{EEGNet}                           & CT & {\ul 65.31} & {\ul 31.14} & 68.32 & 40.60 & {\ul 35.56} & 41.32 & {\ul 40.68} & {\ul 63.54} & {\ul 59.12} & {\ul 49.51}$\pm$1.35 \\ \cmidrule{2-12}
        & FedAvg                  & 64.29 & 30.38 & 59.81 & {\ul 40.63} & 33.36 & 41.23 & 38.05 & 50.43 & 49.13 & 45.26$\pm$1.60               \\
        & FedProx                 & 59.38 & 29.31 & 62.18 & 36.43 & 33.65 & 41.70 & 36.83 & 42.59 & 45.72 & 43.09$\pm$1.59               \\
        & SCAFFOLD                & 36.14 & 28.73 & 36.52 & 30.67 & 30.79 & 32.70 & 33.80 & 31.72 & 38.98 & 33.34$\pm$2.10               \\
        & MOON                    & 63.31 & 28.99 & {\ul 71.30} & 37.96 & 34.49 & 39.44 & 36.08 & 55.01 & 57.23 & 47.09$\pm$1.96               \\
        & FedFA                   & 60.53 & \textbf{34.49} & 60.33 & 39.50 & 33.94 & 41.84 & 40.05 & 47.25 & 52.72 & 45.63$\pm$0.95               \\
        & GA                      & 53.88 & 29.69 & 51.94 & 36.20 & 33.16 & {\ul 42.33} & 31.71 & 38.43 & 46.59 & 40.44$\pm$2.09               \\
        & FedBS                & \textbf{68.52} & 28.96 & \textbf{74.31} & \textbf{44.39} & \textbf{36.69} & \textbf{42.77} & \textbf{49.28} & \textbf{68.11} & \textbf{66.75} & \textbf{53.31}$\pm$0.50               \\ \midrule
        \multirow{8}{*}{DeepConvNet} & CT & {\ul 68.69} & 35.30 & 57.70 & \textbf{42.02} & 31.14 & \textbf{46.15} & 39.12 & {\ul 57.49} & 52.00 & {\ul 47.74}$\pm$0.97 \\ \cmidrule{2-12}
        & FedAvg     & 65.57 & 37.09 & 48.93 & 40.05 & 35.01 & 40.57 & 39.09 & 46.33 & 39.18 & 43.54$\pm$1.37               \\
        & FedProx    & 50.90 & 34.70 & 38.49 & 36.69 & 34.81 & 34.55 & 34.40 & 42.74 & 37.01 & 38.26$\pm$1.38               \\
        & SCAFFOLD   & 37.21 & 27.03 & 29.54 & 26.94 & 27.63 & 29.51 & 27.69 & 30.29 & 32.75 & 29.84$\pm$0.82               \\
        & MOON       & 63.46 & {\ul 36.37} & 46.07 & 39.67 & 33.56 & 42.25 & 37.36 & 47.37 & 38.92 & 42.78$\pm$1.10               \\
        & FedFA      & 64.38 & 34.96 & {\ul 62.24}& 38.77 & 30.70 & 42.25 & {\ul 39.91} & 52.20 & {\ul 52.92} & 46.48$\pm$1.63               \\
        & GA         & 54.74 & \textbf{38.83} & 42.41 & 36.81 & {\ul 35.73} & 36.00 & 33.74 & 42.88 & 35.42 & 39.62$\pm$0.85               \\
        & FedBS   & \textbf{69.41} & 34.15 & \textbf{66.99} & {\ul 41.55} & \textbf{41.15} & {\ul 42.74} & \textbf{49.34} & \textbf{63.57} & \textbf{57.26} & \textbf{51.80}$\pm$1.28               \\ \midrule
        \multirow{8}{*}{ShallowConvNet}                   & CT & 64.79 & 33.10 & \textbf{68.29} & {\ul 39.29} & 34.37 & 43.35 & 42.48 & {\ul 65.45} & 60.85 & {\ul 50.22}$\pm$0.56 \\ \cmidrule{2-12}
        & FedAvg                  & 65.22 & 33.13 & 58.74 & 37.07 & 36.29 & \textbf{46.21} & {\ul 42.88} & 60.62 & 54.05 & 48.24$\pm$1.05               \\
        & FedProx                 & {\ul 66.93} & 32.15 & 60.01 & 37.67 & \textbf{37.15} & \textbf{46.21} & 42.42 & 59.06 & 52.34 & 48.21$\pm$1.25               \\
        & SCAFFOLD                & 62.01 & 28.70 & 65.63 & 35.94 & 31.89 & 32.93 & 39.15 & 61.23 & {\ul 63.28} & 46.75$\pm$1.94               \\
        & MOON                    & 65.54 & {\ul 34.23} & 60.39 & 37.79 & 35.27 & 45.20 & 40.83 & 62.13 & 50.23 & 47.96$\pm$0.79               \\
        & FedFA                   & 50.18 & 33.68 & 44.65 & 35.48 & 33.88 & 37.96 & 41.64 & 50.06 & 49.77 & 41.92$\pm$0.66               \\
        & GA                      & \textbf{67.74} & \textbf{36.08} & 60.33 & 38.83 & 35.50 & {\ul 45.92} & 41.58 & 59.78 & 53.67 & 48.83$\pm$1.49               \\
        & FedBS                & 65.05 & 31.89 & {\ul 67.51} & \textbf{42.10} & {\ul 37.10} & 40.80 & \textbf{46.7}0 & \textbf{65.48} & \textbf{63.46} & \textbf{51.12}$\pm$0.56             \\
        \bottomrule
    \end{tabular}
\end{table*}

\begin{table*}[htbp]    \centering    \setlength{\tabcolsep}{1.6mm}
    \caption{Cross-subject classification accuracies (\%) with EA on MI2. The best accuracy in each column is marked in bold, and the second best by an underline.}
    \label{tab:MI2_EA}
    \begin{tabular}{c|c|cccccccccccccc|c}        \toprule
        \multirow{2}{*}{Model} & \multirow{2}{*}{Approach} & \multicolumn{14}{c|}{Subject}                                                                   & \multirow{2}{*}{Avg.} \\ \cmidrule{3-16}
        &                         & 1    & 2    & 3    & 4    & 5    & 6    & 7    & 8    & 9    & 10   & 11   & 12   & 13   & 14   &                      \\ \midrule
        \multirow{8}{*}{EEGNet}         & CT & 65.0 & 82.5 & {\ul 85.2} & 81.5 & 70.0 & 69.8 & 89.2 & 64.3 & 91.8 & 64.7 & {\ul 80.9} & {\ul 75.2} & 59.2 & \textbf{54.1} & {\ul 73.80}$\pm$0.70 \\ \cmidrule{2-17}
        & FedAvg                  & {\ul 67.4} & 83.4 & 70.8 & 77.9 & 79.7 & 69.1 & 90.2 & 62.0 & 94.1 & {\ul 67.0} & 79.9 & 69.7 & 59.5 & 48.7 & 72.81$\pm$1.10           \\
        & FedProx                 & 64.0 & \textbf{83.9} & 66.5 & {\ul 82.6} & \textbf{81.8} & 67.4 & 88.7 & 55.4 & 94.8 & 65.2 & 80.8 & 69.0 & 57.2 & 50.1 & 71.95$\pm$0.62           \\
        & SCAFFOLD                & 60.5 & 80.4 & 72.3 & 67.2 & 66.4 & 70.1 & \textbf{91.0} & 58.3 & {\ul 94.9} & 66.0 & 75.3 & 65.0 & 59.0 & 50.8 & 69.81$\pm$3.34           \\
        & MOON                    & 64.1 & {\ul 83.5} & 70.4 & \textbf{85.1} & 64.7 & \textbf{73.2} & {\ul 90.9} & 60.6 & \textbf{95.0} & 65.8 & \textbf{81.5} & 75.3 & \textbf{60.9} & 49.4 & 72.89$\pm$2.28           \\
        & FedFA                   & 59.8 & 82.1 & 59.4 & 74.7 & {\ul 81.2} & 70.2 & 86.4 & 64.9 & 92.5 & 66.4 & 78.2 & 67.4 & 58.3 & 49.2 & 70.75$\pm$1.13           \\
        & GA                      & 61.0 & 82.9 & 65.2 & 81.2 & 74.3 & 65.7 & 89.1 & {\ul 68.1} & 93.1 & 63.3 & 80.0 & 67.7 & 55.4 & 49.6 & 71.19$\pm$1.93           \\
        & FedBS                & \textbf{69.5} & 82.9 & \textbf{89.8} & 81.5 & 80.2 & {\ul 72.0} & 88.2 & \textbf{72.4} & 93.1 & \textbf{67.1} & 80.8 & \textbf{81.7} & {\ul 60.7} & {\ul 50.9} & \textbf{76.49}$\pm$0.52          \\ \midrule
        \multirow{8}{*}{DeepConvNet}    & CT & 65.0 & 81.1 & {\ul 85.2} & 83.2 & 76.0 & 71.5 & {\ul 91.7} & 76.3 & 93.0 & {\ul 68.3} & {\ul 80.2} & \textbf{81.7} & {\ul 59.2} & \textbf{55.6} & {\ul 76.28}$\pm$0.91 \\ \cmidrule{2-17}
        & FedAvg                  & 60.1 & 82.8 & 71.8 & 84.3 & \textbf{80.6} & 67.5 & 88.4 & 66.8 & 92.9 & 67.6 & 80.0 & 70.7 & 55.8 & 51.9 & 72.95$\pm$0.37           \\
        & FedProx                 & 57.1 & \textbf{84.1} & 69.4 & {\ul 85.1} & 79.4 & 69.9 & 91.4 & 63.3 & {\ul 93.2} & 67.5 & 76.9 & 67.4 & 55.7 & 51.0 & 72.24$\pm$0.61           \\
        & SCAFFOLD                & 56.4 & 67.8 & 57.4 & 59.5 & 69.8 & 56.9 & 75.9 & 54.3 & 78.9 & 62.6 & 58.6 & 54.6 & 55.3 & 49.4 & 61.23$\pm$1.19           \\
        & MOON                    & 60.3 & {\ul 83.2} & 70.2 & 81.7 & {\ul 80.5} & 68.7 & 88.4 & 64.1 & 92.9 & 67.4 & 76.5 & 69.7 & 56.2 & 50.8 & 72.18$\pm$0.62           \\
        & FedFA                   & {\ul 65.4} & 81.4 & 75.5 & \textbf{85.7} & 77.5 & \textbf{75.8} & \textbf{92.2} & {\ul 82.4} & \textbf{94.0} & \textbf{68.7} & \textbf{80.4} & 71.4 & 58.4 & {\ul 53.8} & 75.89$\pm$1.00           \\
        & GA                      & 56.3 & 81.5 & 72.2 & 76.3 & 79.3 & 66.7 & 89.6 & 63.2 & 92.1 & 67.8 & 77.0 & 68.2 & 55.6 & 51.4 & 71.21$\pm$0.38           \\
        & FedBS                & \textbf{69.4} & 81.8 & \textbf{92.6} & 83.9 & 79.6 & {\ul 71.6} & 87.5 & \textbf{84.6} & 92.6 & 65.7 & 79.1 & {\ul 81.4} & \textbf{59.6} & 52.3 & \textbf{77.25}$\pm$0.52           \\ \midrule
        \multirow{8}{*}{ShallowConvNet} & CT & \textbf{62.0} & \textbf{82.3} & \textbf{86.9} & {\ul 79.2} & {\ul 74.4} & 69.4 & \textbf{89.1} & {\ul 80.3} & 90.3 & 66.8 & {\ul 75.4} & 71.0 & 56.4 & 51.0 & \textbf{73.89}$\pm$0.64 \\ \cmidrule{2-17}
        & FedAvg                  & 59.9 & {\ul 79.2} & 82.0 & 76.7 & 71.3 & 61.3 & 84.5 & 72.4 & 90.6 & 64.9 & 69.5 & 66.8 & 58.0 & 50.8 & 70.55$\pm$0.88           \\
        & FedProx                 & {\ul 61.3} & 79.0 & 82.5 & 74.9 & 70.9 & 63.2 & 83.2 & 72.9 & 90.6 & 66.5 & 69.5 & 64.6 & 56.7 & 52.8 & 70.61$\pm$0.42           \\
        & SCAFFOLD                & 58.8 & 78.7 & {\ul 83.9} & 75.0 & 68.2 & \textbf{71.7} & {\ul 88.4} & 70.2 & \textbf{91.4} & \textbf{69.5} & \textbf{76.6} & {\ul 72.1} & \textbf{59.2} & 51.4 & 72.49$\pm$0.96           \\
        & MOON                    & 60.2 & 79.8 & 82.2 & 74.8 & 71.6 & 61.0 & 86.2 & 71.2 & 90.4 & 64.4 & 70.1 & 65.4 & 57.1 & 52.3 & 70.47$\pm$0.47           \\
        & FedFA                   & 55.9 & 68.7 & 72.5 & 68.1 & 63.4 & 57.9 & 75.0 & 62.3 & 81.8 & 59.7 & 61.5 & 58.5 & 54.1 & {\ul 52.9} & 63.74$\pm$0.94           \\
        & GA                      & 59.5 & 78.9 & 82.8 & 75.3 & 73.4 & 63.8 & 83.1 & 73.8 & 90.6 & 66.7 & 70.8 & 67.9 & {\ul 58.1} & 52.2 & 71.21$\pm$0.96           \\
        & FedBS                & 60.0 & 78.1 & \textbf{86.9} & \textbf{80.0} & \textbf{76.9} & {\ul 70.6} & 83.8 & \textbf{81.9} & {\ul 91.3} & {\ul 66.9} & 73.8 & \textbf{72.5} & 55.0 & \textbf{54.4} & {\ul 73.71}$\pm$0.86           \\        \bottomrule
    \end{tabular}
\end{table*}

\begin{table*}[htbp]    \centering
    \caption{Cross-subject classification accuracies (\%) with EA on MI3. The best accuracy in each column is marked in bold, and the second best by an underline.}
    \label{tab:MI3_EA}
    \begin{tabular}{c|c|cccccccccccc|c}        \toprule
        \multirow{2}{*}{Model} & \multirow{2}{*}{Approach} & \multicolumn{12}{c|}{Subject}                                                     & \multirow{2}{*}{Avg.} \\ \cmidrule{3-14}
        &                         & 1    & 2    & 3    & 4    & 5    & 6    & 7    & 8    & 9    & 10   & 11   & 12   &                      \\ \midrule
        \multirow{8}{*}{EEGNet}         & CT & 89.8 & {\ul 96.1} & 73.1 & {\ul 89.0} & {\ul 83.9} & {\ul 72.3} & {\ul 67.0} & \textbf{64.6} & {\ul 71.9} & 68.9 & {\ul 60.1} & {\ul 55.8} & {\ul 74.37}$\pm$1.20 \\ \cmidrule{2-15}
        & FedAvg                  & 89.0 & 95.2 & 83.0 & \textbf{90.0} & 79.4 & 66.9 & 58.5 & 63.0 & 71.2 & 65.2 & 53.1 & 51.0 & 72.11$\pm$1.67           \\
        & FedProx                 & 92.0 & 95.4 & 83.8 & {\ul 89.0} & 69.0 & 62.0 & 57.2 & 57.3 & 66.6 & 66.5 & 52.0 & 50.8 & 70.13$\pm$0.55           \\
        & SCAFFOLD                & 85.4 & 89.8 & 80.8 & 86.6 & 81.2 & 62.8 & 58.5 & 56.8 & 64.3 & 62.8 & 53.2 & 52.1 & 69.51$\pm$2.11           \\
        & MOON                    & \textbf{96.5} & \textbf{96.3} & 75.3 & 89.5 & 78.3 & 68.8 & 62.2 & 61.2 & 71.2 & \textbf{69.3} & 56.8 & 54.5 & 73.31$\pm$1.90           \\
        & FedFA                   & 92.6 & {\ul 96.1} & 87.5 & 86.0 & 69.8 & 64.2 & 58.2 & 59.1 & 69.5 & 66.9 & 53.8 & 50.5 & 71.18$\pm$1.13           \\
        & GA                      & 84.1 & 93.1 & \textbf{89.3} & 80.3 & 64.5 & 57.1 & 53.3 & 58.4 & 64.1 & 63.3 & 52.1 & 50.3 & 67.49$\pm$0.63           \\
        & FedBS                & {\ul 94.1} & 94.6 & {\ul 88.0} & 87.0 & \textbf{85.8} & \textbf{72.6} & \textbf{69.5} & {\ul 64.5} & \textbf{72.8} & {\ul 69.1} & \textbf{60.6} & \textbf{57.9} & \textbf{76.37}$\pm$0.69           \\ \midrule
        \multirow{8}{*}{DeepConvNet}    & CT & {\ul 96.1} & \textbf{95.5} & 75.7 & {\ul 84.5} & 79.3 & {\ul 66.4} & {\ul 66.5} & {\ul 67.6} & {\ul 72.8} & {\ul 69.9} & \textbf{57.3} & {\ul 53.7} & {\ul  73.78}$\pm$0.47 \\ \cmidrule{2-15}
        & FedAvg                  & 82.4 & 91.6 & 85.0 & 77.5 & 53.4 & 55.7 & 59.9 & 52.6 & 63.1 & 57.1 & 50.6 & 50.0 & 64.92$\pm$0.60           \\
        & FedProx                 & 75.8 & 88.0 & 87.1 & 75.6 & 55.5 & 56.0 & 58.1 & 55.5 & 66.7 & 59.0 & 50.2 & 50.0 & 64.79$\pm$0.75           \\
        & SCAFFOLD                & 66.3 & 73.8 & 77.8 & 67.6 & 60.9 & 56.8 & 57.2 & 51.5 & 61.0 & 54.2 & {\ul 52.8} & 50.0 & 60.83$\pm$0.92           \\
        & MOON                    & 77.8 & 87.9 & \textbf{92.0} & 75.4 & 54.8 & 55.6 & 59.8 & 53.7 & 60.0 & 58.4 & 50.1 & 50.0 & 64.62$\pm$1.37           \\
        & FedFA                   & \textbf{96.9} & {\ul 94.6} & {\ul 90.9} & 80.1 & {\ul 81.8} & 65.4 & 64.4 & 58.9 & 72.5 & 68.7 & 50.8 & 50.3 & 72.93$\pm$0.69           \\
        & GA                      & 77.5 & 87.7 & 89.4 & 72.7 & 55.1 & 54.6 & 64.5 & 53.2 & 60.9 & 61.9 & 50.8 & 50.0 & 64.86$\pm$1.15           \\
        & FedBS                & 91.9 & 93.8 & 87.7 & \textbf{86.3} & \textbf{82.1} & \textbf{74.4} & \textbf{72.0} & \textbf{68.5} & \textbf{75.3} & \textbf{71.4} & 52.6 & \textbf{60.1} & \textbf{76.34}$\pm$0.65           \\ \midrule
        \multirow{8}{*}{ShallowConvNet} & CT & 89.6 & 95.2 & 70.4 & 85.9 & {\ul 82.7} & 69.5 & {\ul 74.3} & 65.0 & \textbf{72.8} & 70.3 & 64.5 & 56.9 & {\ul 74.77}$\pm$0.65 \\ \cmidrule{2-15}
        & FedAvg                  & 89.5 & 93.0 & 74.3 & 86.3 & 77.8 & 68.3 & 73.3 & \textbf{69.3} & 70.8 & 67.3 & 64.8 & 58.5 & 74.40$\pm$0.94           \\
        & FedProx                 & 84.3 & \textbf{96.3} & 71.8 & {\ul 87.3} & 77.8 & {\ul 70.8} & 73.5 & 68.0 & 68.3 & 70.5 & 64.8 & 56.8 & 74.15$\pm$0.84           \\
        & SCAFFOLD                & 71.3 & 91.8 & {\ul 83.5} & 75.4 & 76.5 & 64.1 & 70.3 & 63.0 & 67.3 & 66.3 & 60.8 & 50.8 & 70.10$\pm$0.83           \\
        & MOON                    & 85.8 & {\ul 95.8} & 76.8 & \textbf{88.0} & 77.0 & 66.3 & \textbf{75.0} & 66.3 & 70.3 & 70.3 & 61.8 & 55.8 & 74.06$\pm$1.55           \\
        & FedFA                   & 79.7 & 91.4 & 80.6 & 79.3 & 71.8 & 67.4 & 65.9 & 68.8 & 67.7 & 67.9 & {\ul 66.3} & \textbf{61.0} & 72.32$\pm$0.94           \\
        & GA                      & \textbf{91.5} & 94.3 & 75.8 & 85.5 & 77.0 & 64.0 & 69.8 & {\ul 69.0} & 69.8 & {\ul 70.8} & \textbf{66.8} & 55.5 & 74.13$\pm$1.06           \\
        & FedBS                & {\ul 90.5} & 92.7 & \textbf{86.9} & 81.4 & \textbf{83.3} & \textbf{71.0} & 72.6 & 63.0 & {\ul 71.4} & \textbf{71.3} & 63.3 & {\ul 60.8} & \textbf{75.68}$\pm$0.70       \\   \bottomrule
    \end{tabular}
\end{table*}

\begin{table}[htbp] \centering
	\caption{Adjust $p$-values between FedBS and other approaches. The $p$-values less than 0.05 are marked in bold.}
	\label{tab:p-values}
	\begin{tabular}{c|c|ccc} \toprule
		Model                           & FedBS vs. & MI1    & MI2    & MI3    \\ \midrule
		\multirow{7}{*}{EEGNet}         & CT        & $\boldsymbol{<0.001}$ & $\boldsymbol{<0.001}$ & $\boldsymbol{0.004}$ \\
		& FedAvg    & $\boldsymbol{<0.001}$ & $\boldsymbol{<0.001}$ & $\boldsymbol{<0.001}$ \\
		& FedProx   & $\boldsymbol{<0.001}$ & $\boldsymbol{<0.001}$ & $\boldsymbol{<0.001}$ \\
		& SCAFFOLD  & $\boldsymbol{<0.001}$ & $\boldsymbol{<0.001}$ & $\boldsymbol{<0.001}$ \\
		& MOON      & $\boldsymbol{<0.001}$ & $\boldsymbol{<0.001}$ & $\boldsymbol{<0.001}$ \\
		& FedFA     & $\boldsymbol{<0.001}$ & $\boldsymbol{<0.001}$ & $\boldsymbol{<0.001}$ \\
		& GA        & $\boldsymbol{<0.001}$ & $\boldsymbol{<0.001}$ & $\boldsymbol{<0.001}$ \\ \midrule
		\multirow{7}{*}{DeepConvNet}    & CT        & $\boldsymbol{<0.001}$ & $0.219$ & $\boldsymbol{<0.001}$ \\
		& FedAvg    & $\boldsymbol{<0.001}$ & $\boldsymbol{<0.001}$ & $\boldsymbol{<0.001}$ \\
		& FedProx   & $\boldsymbol{<0.001}$ & $\boldsymbol{<0.001}$ & $\boldsymbol{<0.001}$ \\
		& SCAFFOLD  & $\boldsymbol{<0.001}$ & $\boldsymbol{<0.001}$ & $\boldsymbol{<0.001}$ \\
		& MOON      & $\boldsymbol{<0.001}$ & $\boldsymbol{<0.001}$ & $\boldsymbol{<0.001}$ \\
		& FedFA     & $\boldsymbol{<0.001}$ & $0.063$ & $\boldsymbol{<0.001}$ \\
		& GA        & $\boldsymbol{<0.001}$ & $\boldsymbol{<0.001}$ & $\boldsymbol{<0.001}$ \\ \midrule
		\multirow{7}{*}{ShallowConvNet} & CT        & $0.050$ & $0.313$ & $0.206$ \\
		& FedAvg    & $\boldsymbol{<0.001}$ &$\boldsymbol{<0.001}$ & $0.059$ \\
		& FedProx   & $\boldsymbol{<0.001}$ & $\boldsymbol{<0.001}$ & $0.082$ \\
		& SCAFFOLD  & $\boldsymbol{<0.001}$ & $0.092$ & $\boldsymbol{<0.001}$ \\
		& MOON      & $\boldsymbol{<0.001}$ & $\boldsymbol{<0.001}$ & $0.115$ \\
		& FedFA     & $\boldsymbol{<0.001}$ & $\boldsymbol{<0.001}$ & $\boldsymbol{<0.001}$ \\
		& GA        & $\boldsymbol{0.005}$ & $\boldsymbol{0.008}$ & $0.060$ \\ \bottomrule
	\end{tabular}
\end{table}

\subsection{Ablation Studies}

We performed ablation studies to confirm the effectiveness of the two components (local batch-specific BN, and SAM) in FedBS. We also performed paired $t$-tests on the ablation studies results, and adjusted the $p$-values using Benjamini-Hochberg False Discovery Rate correction. The results are shown in Table~\ref{tab:ablation}, where $\star$ indicates that the adjusted $p$-value of the paired $t$-test between FedBS and a variant is less than 0.05. Further details of the paired $t$-tests are provided in Appendix.

\begin{table}[htbp]    \centering
    \caption{Average classification accuracies (\%) in ablation studies. The best accuracy in each column is marked in bold. Adjusted $p$-values of the paired $t$-tests between FedBS (with local batch-specific BN and SAM) and others less than 0.05 are indicated with $\star$. Note that the 'Avg.' column did not participate in the paired $t$-tests.}    \label{tab:ablation}
    \begin{tabular}{c|cc|ccc|c}        \toprule
        \multirow{2}{*}{Model}          & \multicolumn{2}{c|}{Strategy}                             & \multicolumn{3}{c|}{Datasets} & \multirow{2}{*}{Avg.} \\ \cmidrule{2-6}
        & BN                            & SAM                         & MI1   & MI2   & MI3   &       \\ \midrule
        \multirow{4}{*}{EEGNet}         & \XSolidBrush & \XSolidBrush & 45.26$^\star$    & 72.81$^\star$    & 72.11$^\star$    & 63.63                \\
        & \Checkmark   & \XSolidBrush   & 51.89$^\star$ & 75.97 & 76.08 & 67.98 \\
        & \XSolidBrush & \Checkmark     & 50.14$^\star$ & 75.32$^\star$ & 74.79$^\star$ & 66.75 \\
        & \Checkmark   & \Checkmark     & \textbf{53.31} & \textbf{76.49} & \textbf{76.37} & \textbf{68.72} \\ \midrule
        \multirow{4}{*}{DeepConvNet}    & \XSolidBrush & \XSolidBrush & 43.54$^\star$    & 72.95$^\star$    & 64.92$^\star$    & 60.47                \\
        & \Checkmark   & \XSolidBrush   & 50.37$^\star$ & 76.24$^\star$ & 75.83 & 67.48 \\
        & \XSolidBrush & \Checkmark         & 46.28$^\star$ & 73.88$^\star$ & 66.94$^\star$ & 62.37 \\
        & \Checkmark   & \Checkmark     & \textbf{51.80} & \textbf{77.25} & \textbf{76.34} & \textbf{68.46} \\ \midrule
        \multirow{4}{*}{ShallowConvNet} & \XSolidBrush & \XSolidBrush & 48.24$^\star$    & 70.55$^\star$    & 74.40    & 64.40                \\
        & \Checkmark   & \XSolidBrush   & 49.59$^\star$ & 71.04$^\star$ & 75.16 & 65.26 \\
        & \XSolidBrush & \Checkmark         & 49.73$^\star$ & 72.43 & 75.41 & 65.86 \\
        & \Checkmark   & \Checkmark     & \textbf{51.12} & \textbf{72.68} & \textbf{75.68} & \textbf{66.49} \\        \bottomrule
    \end{tabular}
\end{table}

Table~\ref{tab:ablation} shows that every individual strategy was effective, and their combination achieved the best performance in all scenarios. Specifically, using only the local batch-specific BN, the three models improved 4.35\%, 7.01\% and 1.23\% respectively over FedAvg. Using only the SAM, the three models improved 3.12\%, 1.90\% and 1.33\% respectively over FedAvg. When the two strategies were combined, the three models improved 5.09\%, 7.99\% and 2.09\% respectively over FedAvg.

To emphasize again, local batch-specific BN enhances generalization by making the feature distributions more uniform and adaptable to new subjects, and SAM improves generalization by introducing perturbations during gradient computation.

\subsection{Effect of FL parameters}

\begin{figure*}[htbp]	\centering
	\subfigure[MI1]{\label{fig:paremeter_MI1_PK}   \includegraphics[width=0.32\linewidth]{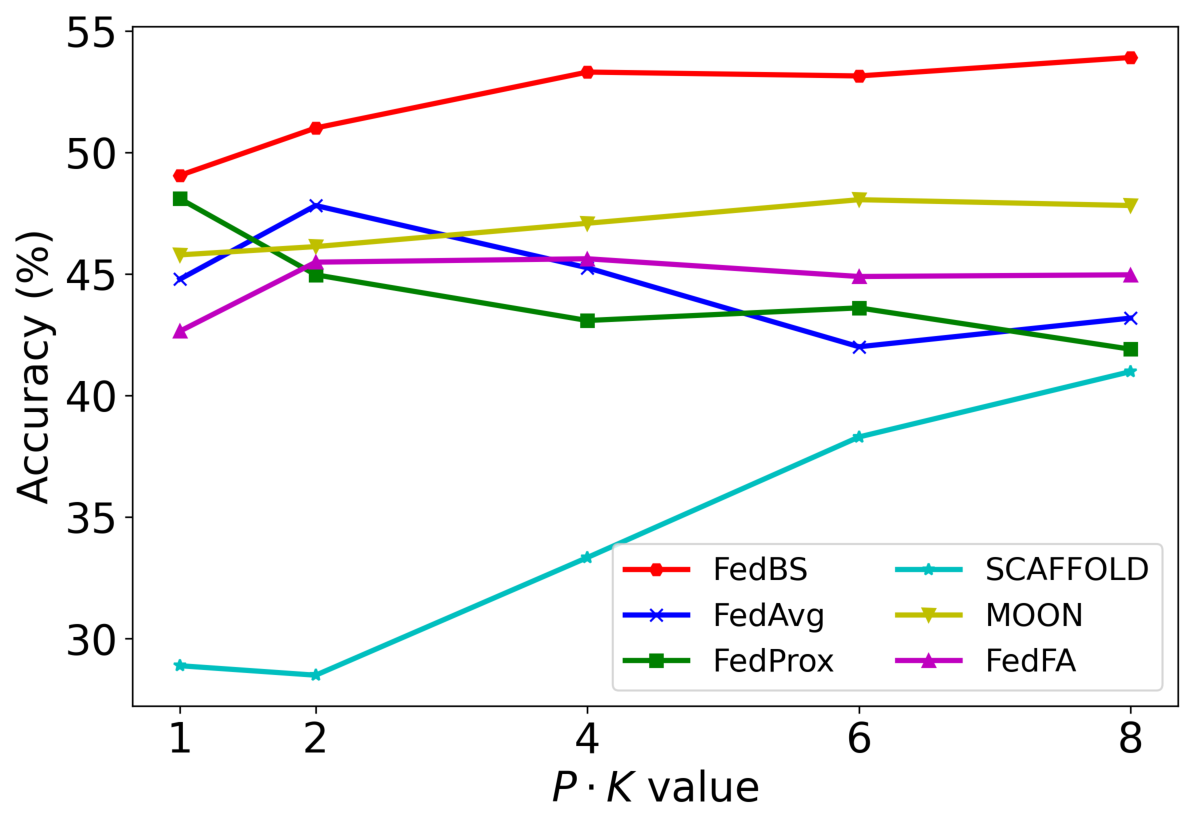}}
	\subfigure[MI2]{\label{fig:paremeter_MI2_PK}   \includegraphics[width=0.32\linewidth]{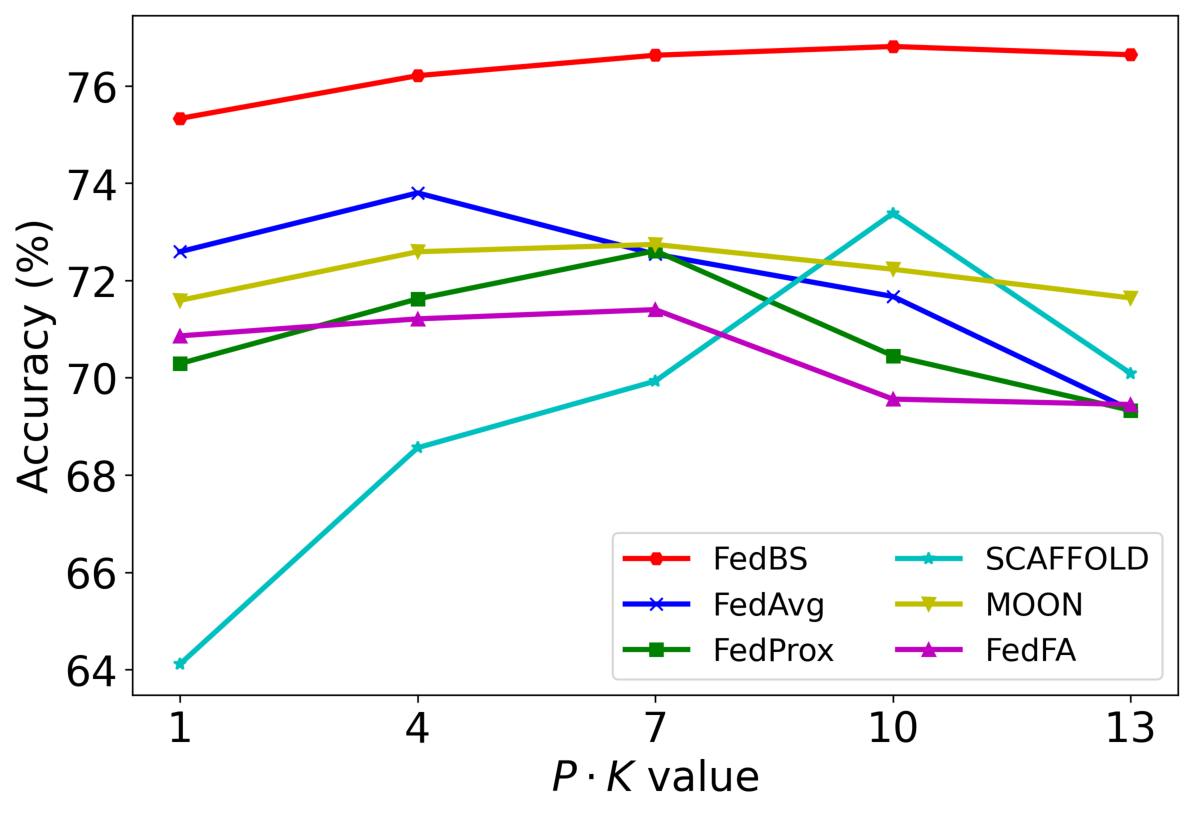}}
	\subfigure[MI3]{\label{fig:paremeter_MI3_PK}   \includegraphics[width=0.32\linewidth]{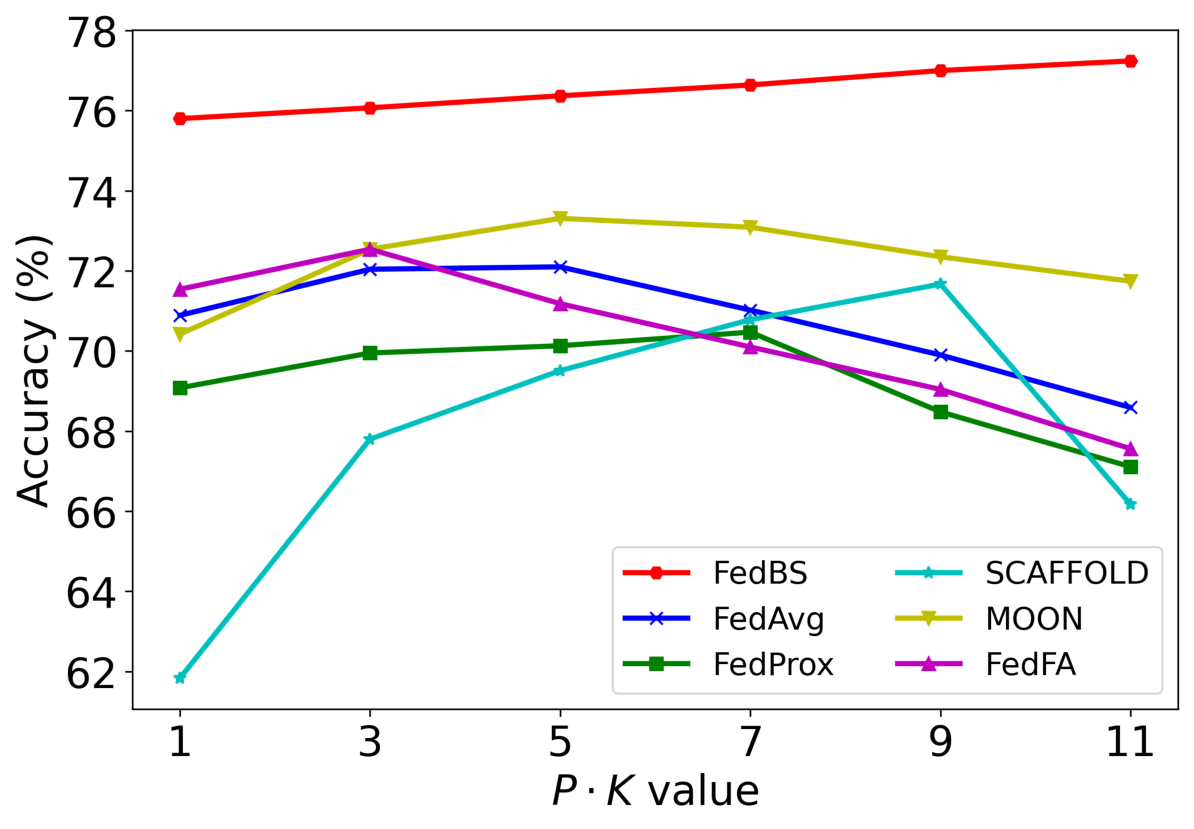}}
	\caption{Average classification accuracies of different FL approaches w.r.t. $P\cdot K$, the number of selected clients. GA was not included because it requires the participation of all clients in each round of training. (a) MI1; (b) MI2; and, (c) MI3.}
	\label{fig:prameter_PK}
\end{figure*}

\begin{figure*}[htbp]	\centering
	\subfigure[MI1]{\label{fig:paremeter_MI1_E}   \includegraphics[width=0.32\linewidth]{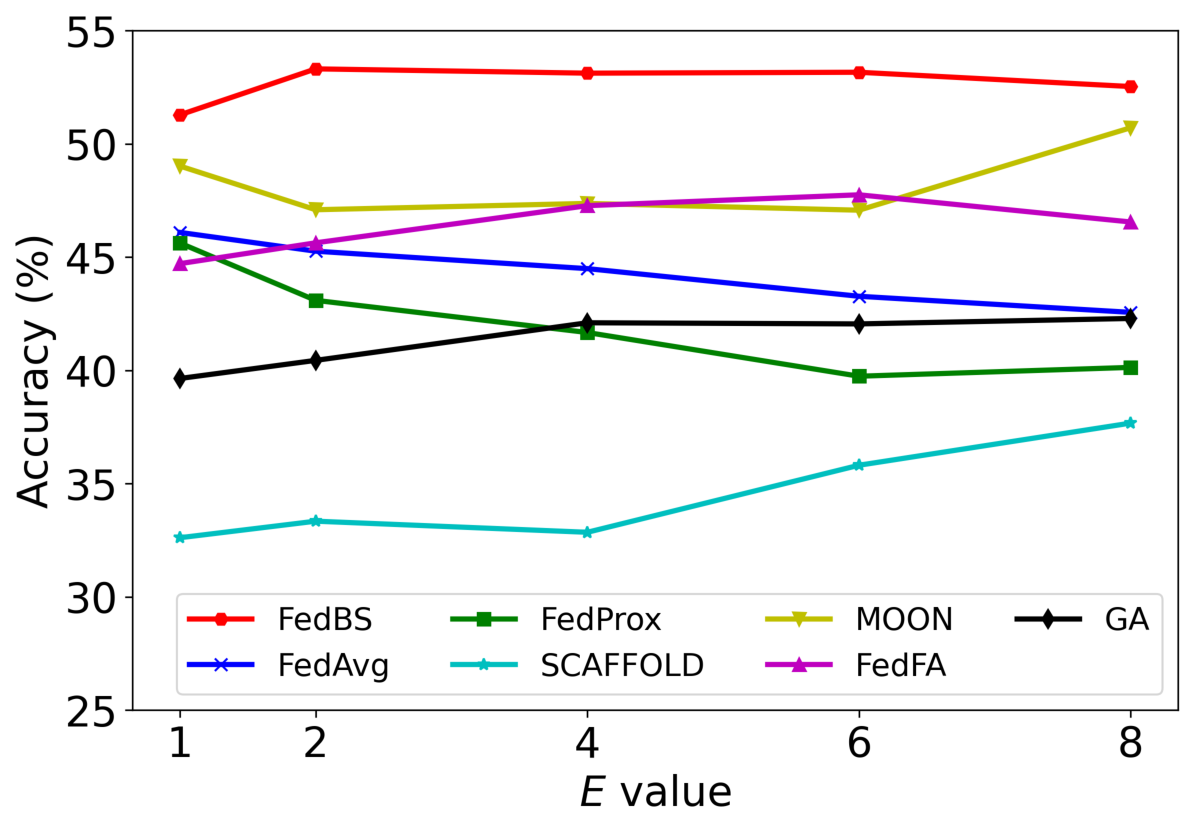}}
	\subfigure[MI2]{\label{fig:paremeter_MI2_E}   \includegraphics[width=0.32\linewidth]{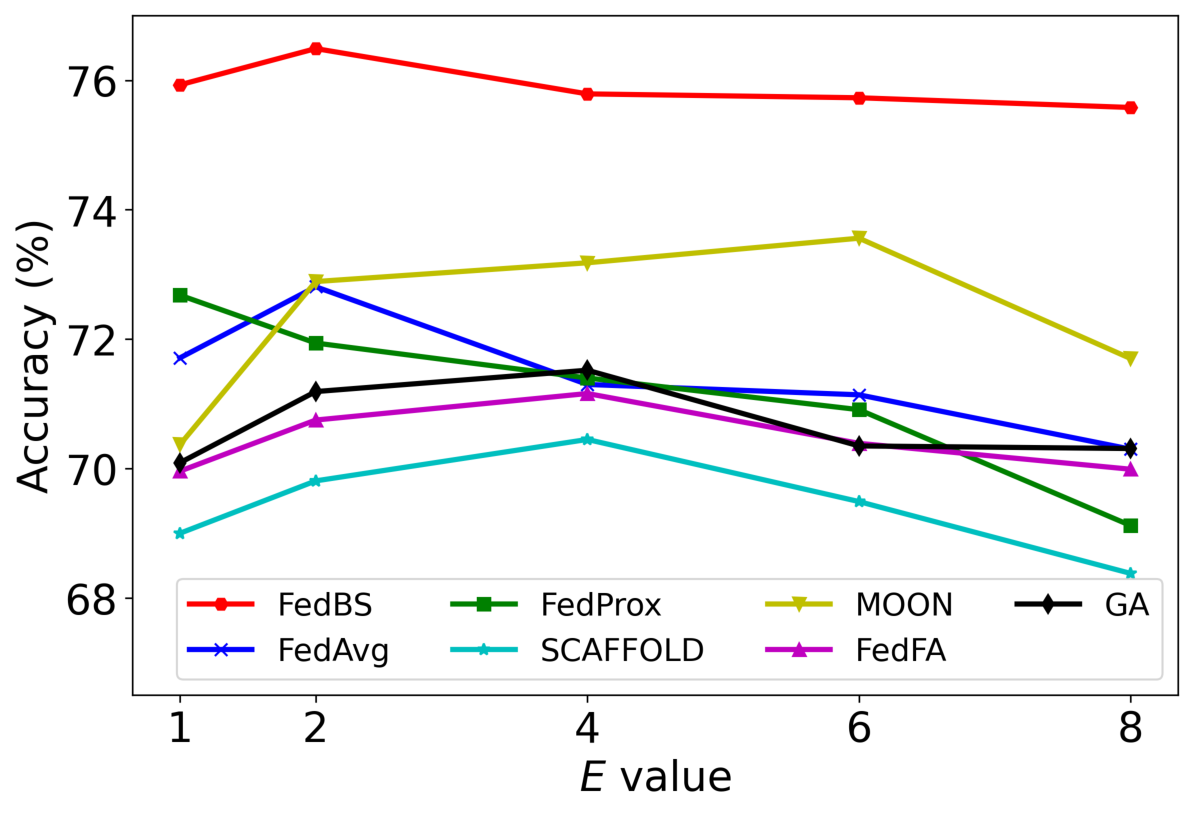}}
	\subfigure[MI3]{\label{fig:paremeter_MI3_E}   \includegraphics[width=0.32\linewidth]{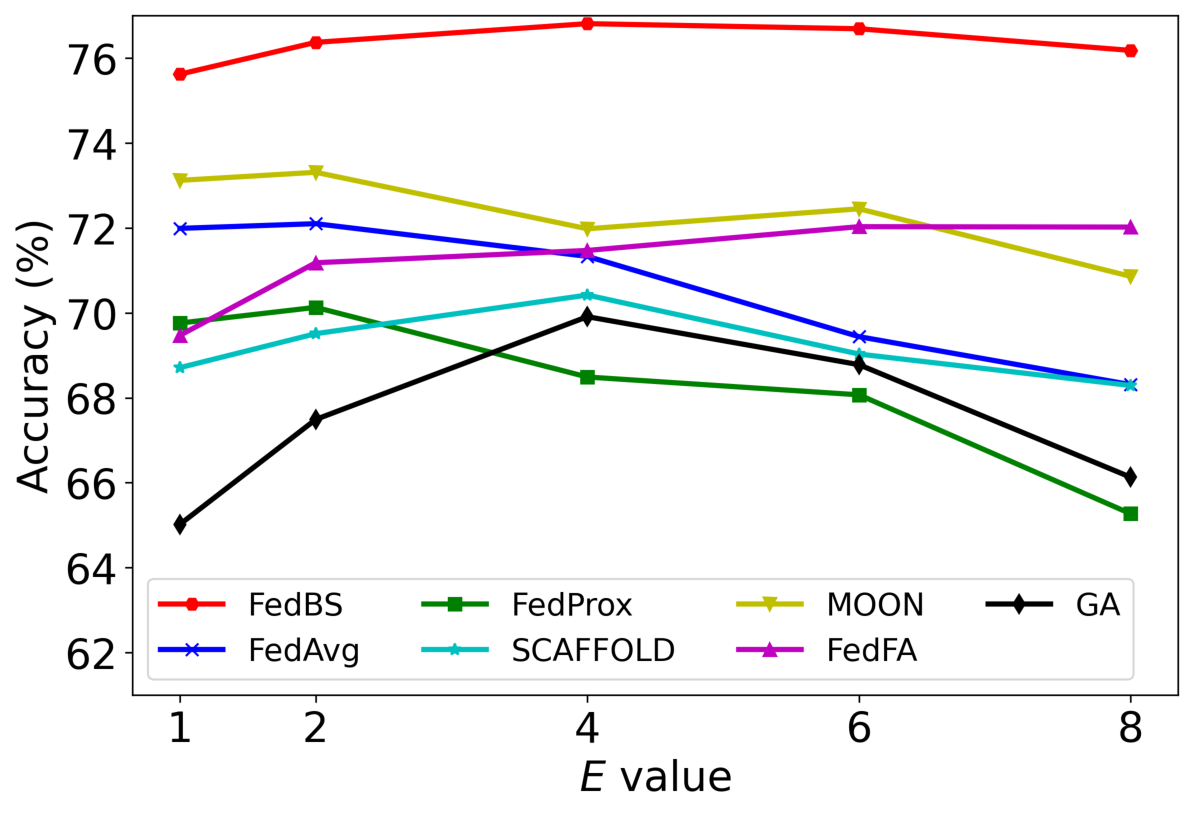}}
	\caption{Average classification accuracies of different FL approaches w.r.t. $E$, the number of local computation epochs. (a) MI1; (b) MI2; and, (c) MI3.}
	\label{fig:prameter_E}
\end{figure*}

Figs.~\ref{fig:prameter_PK} and \ref{fig:prameter_E} show the performance of different FL approaches with varying client selection weight $P$ and number of local computation epochs $E$, respectively, when EEGNet was used as the backbone. Note that for easier understanding, the horizontal axis of Fig.~\ref{fig:prameter_PK} is $P\cdot K$, the number of selected clients.

FedBS always outperformed other FL approaches, regardless of $P$ and $E$. Particularly, Fig.~\ref{fig:prameter_PK} shows that FedBS's performance gradually increased with $P$, whereas other approaches did not have a consistent pattern.

In practice, one must carefully select the appropriate $P$ and $E$ to trade-off the communication cost and the classification accuracy. However, no matter which $P$ and $E$ are used, our proposed FedBS always had the best performance.

\subsection{Visualization}

\begin{figure*}[htbp]	\centering
	\subfigure[CT]{\label{fig:tsne_cen}   \includegraphics[width=0.32\linewidth]{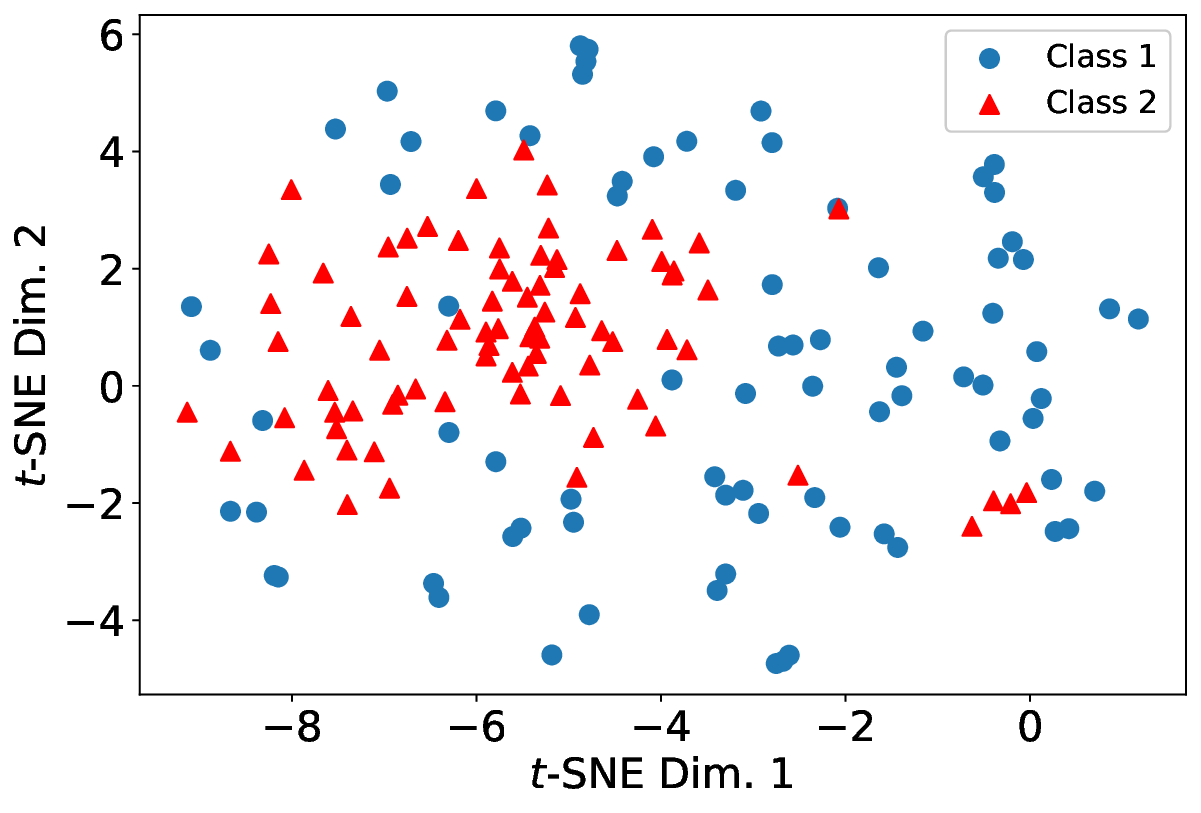}}
	\subfigure[FedAvg]{\label{fig:tsne_fedavg}   \includegraphics[width=0.32\linewidth]{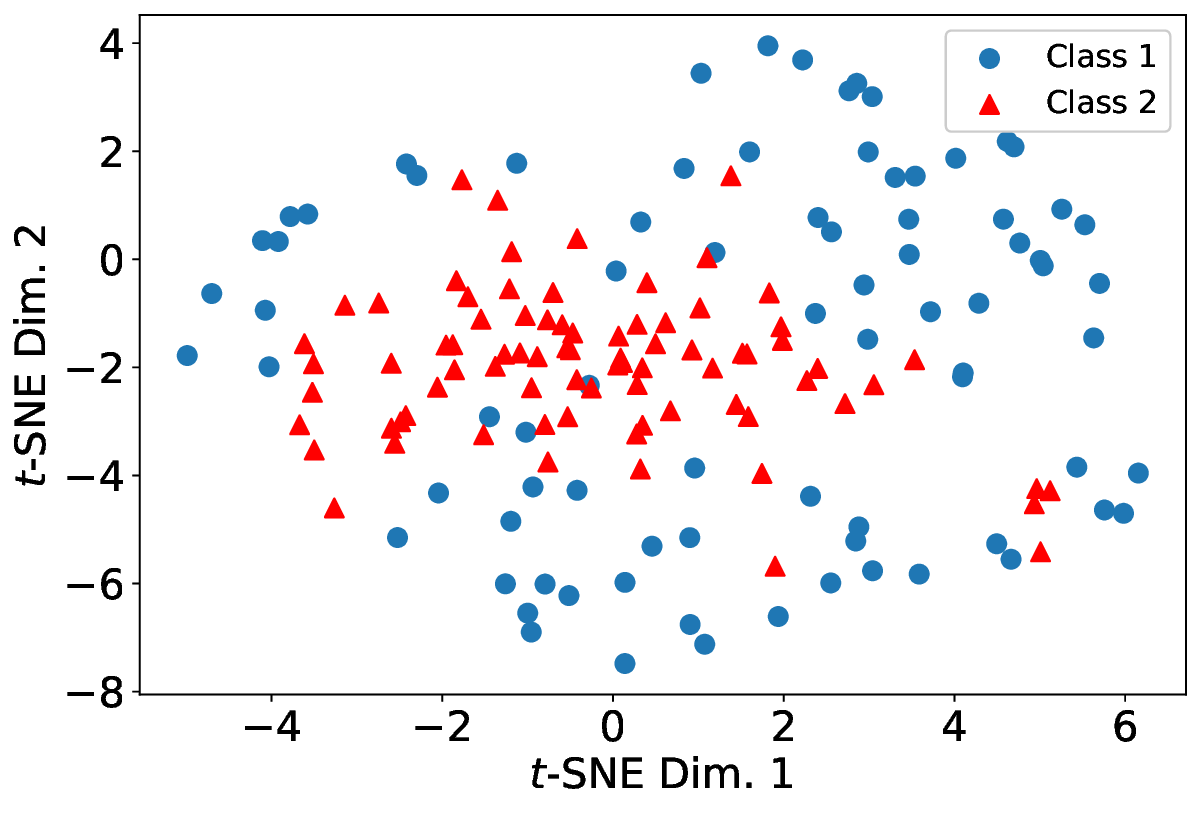}}
	\subfigure[FedBS]{\label{fig:tsne_fedbs}   \includegraphics[width=0.32\linewidth]{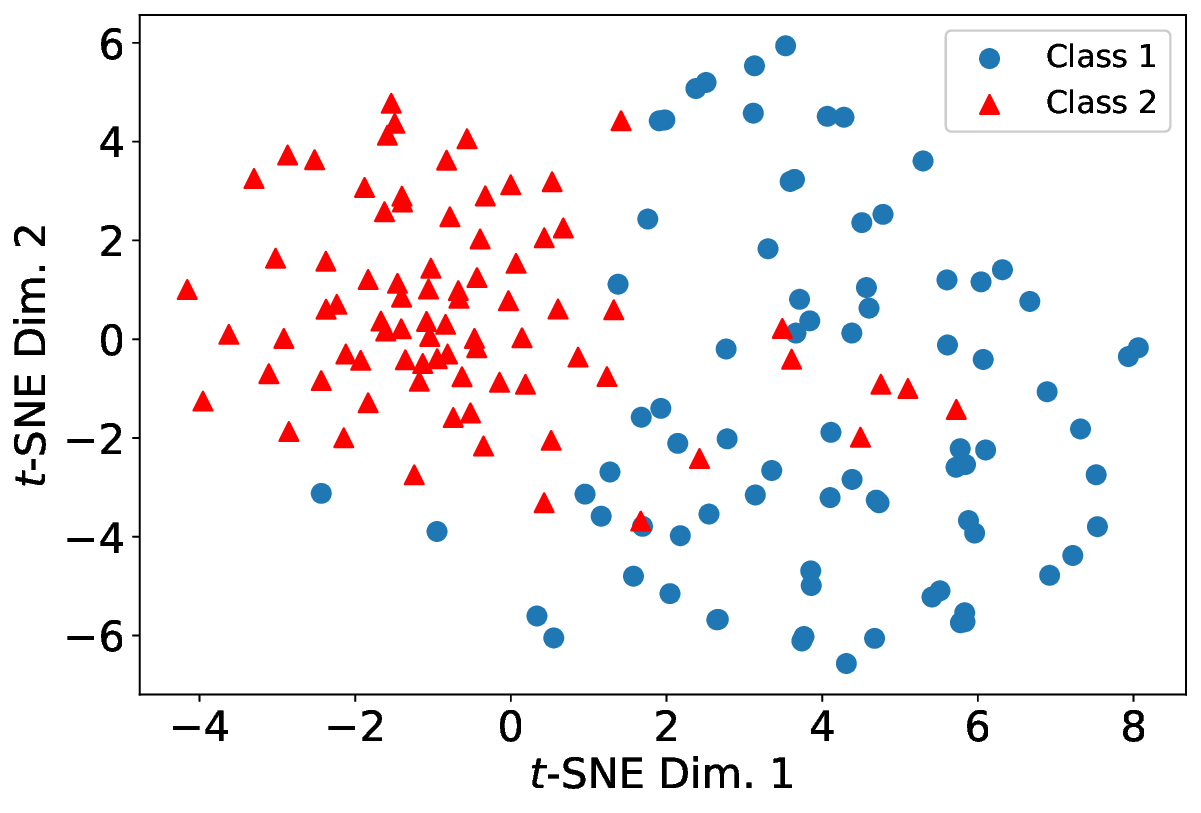}}
	\caption{$t$-SNE visualization of feature extracted from test Subject 1 on the MI2 dataset. (a) CT; (b) FedAvg; and, (c) FedBS.}
	\label{fig:tsne}
\end{figure*}

Fig.~\ref{fig:tsne} shows $t$-SNE~\cite{Van2008} visualizations of features extracted by CT, FedAvg and FedBS from test Subject 1 on the MI2 dataset. Table~\ref{tab:gdv} presents the average Generalized Discrimination Values \cite{schilling2021quantifying} calculated on features extracted by CT, FedAvg and FedBS from each test subject in the three datasets. This metric, ranging from $[-1, 0]$, quantifies the separability of neural network features, with lower values indicating better separability. Manhattan distance was used in the calculation as it is more suitable for high-dimensional data \cite{Aggarwal2001}. Clearly, FedBS's features were more separable. This is because FedBS utilizes local batch-specific BN to align samples from different subjects, reducing the distribution disparities and improving the classification performance on new subjects.

\begin{table}[htbp]	\centering
	\caption{The average Generalized Discrimination Values calculated on features extracted by CT, FedAvg, and FedBS from each test subject in three datasets. The best value in each column is marked in bold.}
	\label{tab:gdv}
	\begin{tabular}{c|ccc|c}
		\toprule
		\multicolumn{1}{c|}{\multirow{2}{*}{Approach}} & \multicolumn{3}{c|}{Datasets} & \multicolumn{1}{c}{\multirow{2}{*}{Avg.}} \\ \cmidrule{2-4}
		\multicolumn{1}{c|}{}                          & MI1      & MI2      & MI3     & \multicolumn{1}{c}{}                      \\ \midrule
		CT                                             & -0.3397  & -0.3479  & -0.3755 & -0.3544                                   \\
		FedAvg                                         & -0.2941  & -0.3294  & -0.3167 & -0.3134                                   \\
		FedBS                                          & \textbf{-0.3964}  & \textbf{-0.3771}  & \textbf{-0.3932} & \textbf{-0.3889}                                   \\ \bottomrule
	\end{tabular}
\end{table}

\subsection{Effect of Test Batch Size}

\begin{figure}[htbp]	\centering
	\includegraphics[width=0.8\linewidth]{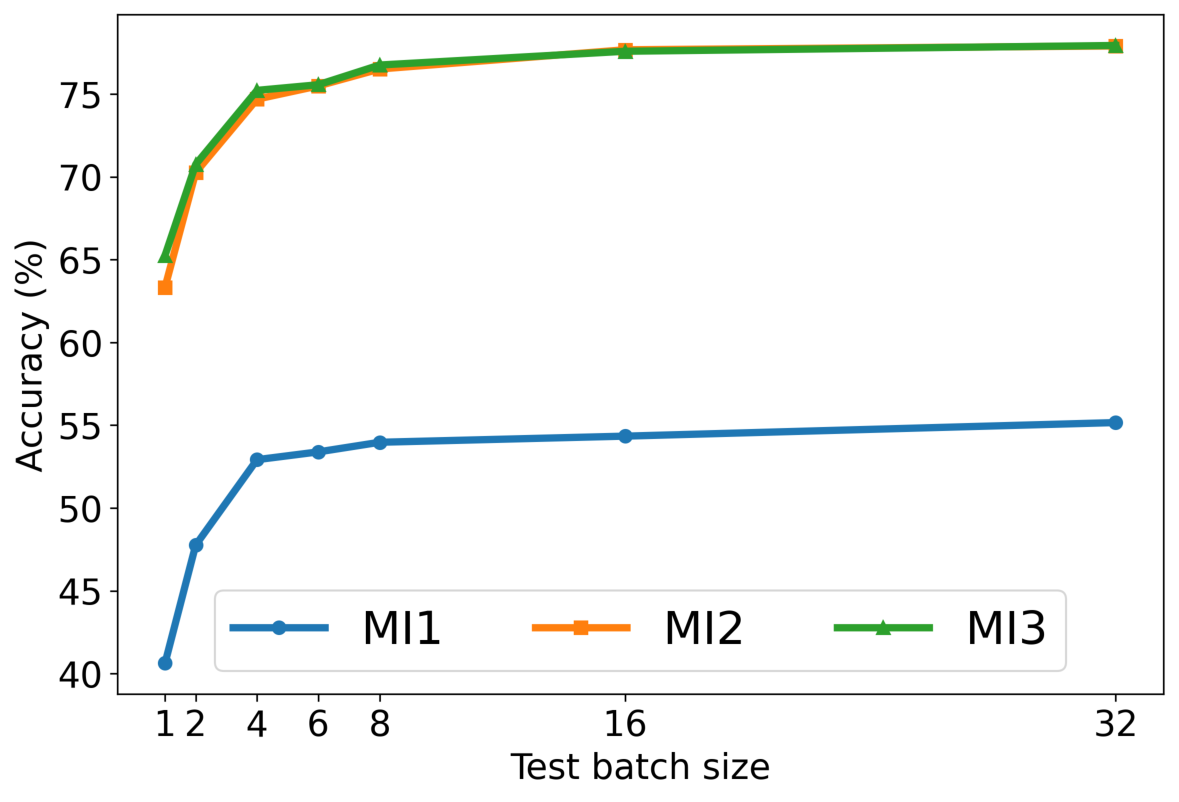}
	\caption{Classification accuracies of FedBS under different test batch sizes.}	\label{fig:parameter_batch}
\end{figure}

As FedBS calculates the BN statistics for each batch, the batch size also impacts the test results. Fig.~\ref{fig:parameter_batch} shows the performance of FedBS under different test batch sizes, when EEGNet was used. The performance increased with the test batch size, but converged at 4.

\section{Discussions and Future Research} \label{sect:disscussion}

The experimental results in the previous section demonstrated that both the local batch-specific BN and the SAM optimizer introduced by FedBS effectively enhance the cross-subject decoding accuracy. In FL, local batch-specific BN samples data from individual subjects, allowing for subject-specific normalization. Models trained individually on subjects in FL are biased and sensitive to perturbations due to limited training data, making them suitable for the SAM optimizer. However, these conditions do not hold in CT; so, how to extend local batch-specific BN and the SAM optimizer to CT requires further research.

Our current work also has some limitations, e.g., FedBS was only validated in the classic MI-based BCIs, whereas there are many other popular BCI paradigms; and, FedBS only considers the simple homogeneous EEG classification scenario, where the number and locations of EEG channels from all subjects are the same. To expand the applicability of FedBS, our future research will:
\begin{enumerate}
\item Extend FedBS to other BCI paradigms, e.g., steady-state visual evoked potentials, P300 event related potentials, and affective BCIs \cite{drwuPIEEE2023}.
\item Extend FedBS to more challenging and more flexible heterogeneous BCI applications, where the number and/or locations of EEG channels are different for different subjects.
\end{enumerate}

\section{Conclusions} \label{sect:conclusion}

This paper has proposed FedBS for privacy protection in EEG-based MI classification. FedBS utilizes local batch-specific BN to reduce data discrepancies among different clients, and SAM optimizer in local training to improve model generalization. Experiments on three public MI datasets using three popular deep learning models demonstrated that FedBS outperformed six state-of-the-art FL approaches. Remarkably, it also outperformed centralized training, which does not consider privacy protection at all.

FedBS protects user EEG data privacy, enabling multiple BCI users to participate in large-scale machine learning model training, which in turn improves the BCI decoding accuracy. We believe it is valuable in facilitating more real-world applications of BCIs.



\appendix

\subsection{Paired $t$-tests between FedBS and other approaches}

We conducted paired $t$-tests between FedBS and other approaches. The performance variable was test accuracy, the test-statistic value was the $t$-value, the effect size was Cohen's $d$, and the confidence level was 95\%.

Tables~\ref{tab:MI1_ttest}-\ref{tab:MI3_ttest} present the paired $t$-test details between FedBS and other approaches on MI1, MI2 and MI3, respectively, including sample means, standard errors, Cohen's $d$ values, $t$-values, and $p$-values. The degrees of freedom for these tests were $53$, $83$, and $71$, respectively.

\begin{table*}[htbp] \centering
	\caption{Paired $t$-test details between FedBS and other approaches on MI1. $p$-values smaller than 0.05 are marked in bold.}
	\label{tab:MI1_ttest}
	\begin{tabular}{c|c|cc|cc|ccc}
		\toprule
		Model                           & FedBS vs. & Mean (FedBS)            & SE (FedBS)             & Mean (Compared) & SE (Compared) & Cohen's $d$ & $t$-value & $p$-value \\ \midrule
		\multirow{7}{*}{EEGNet}         & CT        & \multirow{7}{*}{53.308} & \multirow{7}{*}{2.153} & 49.508          & 1.897         & 0.820     & 6.023   & $\boldsymbol{<0.001}$   \\
		& FedAvg   &  &  & 45.257 & 1.616 & 0.945 & 6.946  & $\boldsymbol{<0.001}$ \\
		& FedProx  &  &  & 43.088 & 1.588 & 1.021 & 7.501  & $\boldsymbol{<0.001}$ \\
		& SCAFFOLD &  &  & 33.337 & 0.938 & 1.352 & 9.932  & $\boldsymbol{<0.001}$ \\
		& MOON     &  &  & 47.090 & 2.076 & 0.775 & 5.692  & $\boldsymbol{<0.001}$ \\
		& FedFA    &  &  & 45.627 & 1.382 & 0.886 & 6.509  & $\boldsymbol{<0.001}$ \\
		& GA       &  &  & 40.436 & 1.233 & 1.167 & 8.575  & $\boldsymbol{<0.001}$ \\ \midrule
		\multirow{7}{*}{DeepConvNet}    & CT        & \multirow{7}{*}{51.794} & \multirow{7}{*}{1.725} & 47.734          & 1.606         & 0.635     & 4.665   & $\boldsymbol{<0.001}$   \\
		& FedAvg   &  &  & 43.535 & 1.277 & 0.944 & 6.937  & $\boldsymbol{<0.001}$ \\
		& FedProx  &  &  & 38.253 & 0.853 & 1.345 & 9.886  & $\boldsymbol{<0.001}$ \\
		& SCAFFOLD &  &  & 29.845 & 0.633 & 1.994 & 14.656 & $\boldsymbol{<0.001}$ \\
		& MOON     &  &  & 42.779 & 1.224 & 1.011 & 7.431  & $\boldsymbol{<0.001}$ \\
		& FedFA    &  &  & 46.480 & 1.606 & 0.876 & 6.435  & $\boldsymbol{<0.001}$ \\
		& GA       &  &  & 39.618 & 0.893 & 1.223 & 8.989  & $\boldsymbol{<0.001}$ \\ \midrule
		\multirow{7}{*}{ShallowConvNet} & CT        & \multirow{7}{*}{51.119} & \multirow{7}{*}{1.850} & 50.219          & 1.877         & 0.273     & 2.005   & $0.050$   \\
		& FedAvg   &  &  & 48.245 & 1.566 & 0.515 & 3.781  & $\boldsymbol{<0.001}$ \\
		& FedProx  &  &  & 48.215 & 1.588 & 0.505 & 3.711  & $\boldsymbol{<0.001}$ \\
		& SCAFFOLD &  &  & 46.750 & 2.144 & 0.787 & 5.786  & $\boldsymbol{<0.001}$ \\
		& MOON     &  &  & 47.955 & 1.620 & 0.504 & 3.702  & $\boldsymbol{<0.001}$ \\
		& FedFA    &  &  & 41.921 & 0.955 & 1.134 & 8.333  & $\boldsymbol{<0.001}$ \\
		& GA       &  &  & 48.826 & 1.587 & 0.396 & 2.913  & $\boldsymbol{0.005}$ \\ \bottomrule
	\end{tabular}
\end{table*}

\begin{table*}[htbp] \centering
	\caption{Paired $t$-test details between FedBS and other approaches on MI2. $p$-values smaller than 0.05 are marked in bold.}
	\label{tab:MI2_ttest}
	\begin{tabular}{c|c|cc|cc|ccc}
		\toprule
		Model                           & FedBS vs. & Mean (FedBS)            & SE (FedBS)             & Mean (Compared) & SE (Compared) & Cohen's $d$ & $t$-value & $p$-value \\ \midrule
		\multirow{7}{*}{EEGNet}         & CT        & \multirow{7}{*}{76.488} & \multirow{7}{*}{1.260} & 73.802          & 1.271         & 0.564     & 5.165   & $\boldsymbol{<0.001}$   \\
		& FedAvg   &  &  & 72.805 & 1.354 & 0.535 & 4.906  & $\boldsymbol{<0.001}$ \\
		& FedProx  &  &  & 71.943 & 1.502 & 0.519 & 4.757  & $\boldsymbol{<0.001}$ \\
		& SCAFFOLD &  &  & 69.806 & 1.513 & 0.661 & 6.054  & $\boldsymbol{<0.001}$ \\
		& MOON     &  &  & 72.894 & 1.508 & 0.386 & 3.538  & $\boldsymbol{<0.001}$ \\
		& FedFA    &  &  & 70.752 & 1.347 & 0.649 & 5.948  & $\boldsymbol{<0.001}$ \\
		& GA       &  &  & 71.190 & 1.410 & 0.658 & 6.029  & $\boldsymbol{<0.001}$ \\ \midrule
		\multirow{7}{*}{DeepConvNet}    & CT        & \multirow{7}{*}{77.247} & \multirow{7}{*}{1.287} & 76.282          & 1.273         & 0.135     & 1.239   & $0.219$   \\
		& FedAvg   &  &  & 72.947 & 1.331 & 0.549 & 5.031  & $\boldsymbol{<0.001}$ \\
		& FedProx  &  &  & 72.239 & 1.427 & 0.543 & 4.978  & $\boldsymbol{<0.001}$ \\
		& SCAFFOLD &  &  & 61.228 & 1.033 & 1.458 & 13.363 & $\boldsymbol{<0.001}$ \\
		& MOON     &  &  & 72.180 & 1.332 & 0.609 & 5.581  & $\boldsymbol{<0.001}$ \\
		& FedFA    &  &  & 75.893 & 1.263 & 0.206 & 1.885  & $0.063$ \\
		& GA       &  &  & 71.212 & 1.330 & 0.734 & 6.725  & $\boldsymbol{<0.001}$ \\ \midrule
		\multirow{7}{*}{ShallowConvNet} & CT        & \multirow{7}{*}{73.715} & \multirow{7}{*}{1.187} & 73.884          & 1.293         & -0.111     & -1.015   & $0.313$   \\
		& FedAvg   &  &  & 70.551 & 1.222 & 0.414 & 3.791  & $\boldsymbol{<0.001}$ \\
		& FedProx  &  &  & 70.611 & 1.177 & 0.407 & 3.732  & $\boldsymbol{<0.001}$ \\
		& SCAFFOLD &  &  & 72.485 & 1.247 & 0.186 & 1.705  & $0.092$ \\
		& MOON     &  &  & 70.469 & 1.230 & 0.408 & 3.735  & $\boldsymbol{<0.001}$ \\
		& FedFA    &  &  & 63.735 & 0.952 & 1.529 & 14.015 & $\boldsymbol{<0.001}$ \\
		& GA       &  &  & 71.205 & 1.177 & 0.295 & 2.706  & $\boldsymbol{0.008}$ \\ \bottomrule
	\end{tabular}
\end{table*}

\begin{table*}[htbp] \centering
	\caption{Paired $t$-test details between FedBS and other approaches on MI3. $p$-values smaller than 0.05 are marked in bold.}
	\label{tab:MI3_ttest}
	\begin{tabular}{c|c|cc|cc|ccc}
		\toprule
		Model                           & FedBS vs. & Mean (FedBS)            & SE (FedBS)             & Mean (Compared) & SE (Compared) & Cohen's $d$ & $t$-value & $p$-value \\ \midrule
		\multirow{7}{*}{EEGNet}         & CT        & \multirow{7}{*}{76.375} & \multirow{7}{*}{1.484} & 74.368          & 1.492         & 0.349     & 2.960   & $\boldsymbol{0.004}$   \\
		& FedAvg   &  &  & 72.104 & 1.789 & 0.685 & 5.812  & $\boldsymbol{<0.001}$ \\
		& FedProx  &  &  & 70.129 & 1.883 & 0.914 & 7.755  & $\boldsymbol{<0.001}$ \\
		& SCAFFOLD &  &  & 69.507 & 1.774 & 0.932 & 7.905  & $\boldsymbol{<0.001}$ \\
		& MOON     &  &  & 73.306 & 1.757 & 0.445 & 3.779  & $\boldsymbol{<0.001}$ \\
		& FedFA    &  &  & 71.181 & 1.822 & 0.851 & 7.219  & $\boldsymbol{<0.001}$ \\
		& GA       &  &  & 67.490 & 1.804 & 1.139 & 9.667  & $\boldsymbol{<0.001}$ \\ \midrule
		\multirow{7}{*}{DeepConvNet}    & CT        & \multirow{7}{*}{76.340} & \multirow{7}{*}{1.449} & 73.781          & 1.527         & 0.477     & 4.045   & $\boldsymbol{<0.001}$   \\
		& FedAvg   &  &  & 64.917 & 1.737 & 1.383 & 11.734 & $\boldsymbol{<0.001}$ \\
		& FedProx  &  &  & 64.788 & 1.618 & 1.399 & 11.868 & $\boldsymbol{<0.001}$ \\
		& SCAFFOLD &  &  & 60.826 & 1.167 & 1.868 & 15.849 & $\boldsymbol{<0.001}$ \\
		& MOON     &  &  & 64.618 & 1.742 & 1.304 & 11.067 & $\boldsymbol{<0.001}$ \\
		& FedFA    &  &  & 72.927 & 1.852 & 0.577 & 4.896  & $\boldsymbol{<0.001}$ \\
		& GA       &  &  & 64.858 & 1.620 & 1.370 & 11.627 & $\boldsymbol{<0.001}$ \\ \midrule
		\multirow{7}{*}{ShallowConvNet} & CT        & \multirow{7}{*}{75.687} & \multirow{7}{*}{1.313} & 74.766          & 1.304         & 0.150     & 1.276   & $0.206$   \\
		& FedAvg   &  &  & 74.404 & 1.154 & 0.226 & 1.919  & $0.059$ \\
		& FedProx  &  &  & 74.158 & 1.211 & 0.208 & 1.764  & $0.082$ \\
		& SCAFFOLD &  &  & 70.097 & 1.289 & 0.843 & 7.149  & $\boldsymbol{<0.001}$ \\
		& MOON     &  &  & 74.063 & 1.253 & 0.188 & 1.596  & $0.115$ \\
		& FedFA    &  &  & 72.316 & 1.010 & 0.517 & 4.389  & $\boldsymbol{<0.001}$ \\
		& GA       &  &  & 74.137 & 1.183 & 0.225 & 1.911  & $0.060$ \\ \bottomrule
	\end{tabular}
\end{table*}

\subsection{Paired $t$-Tests for Ablation Studies}

We conducted paired $t$-tests for the ablation studies. The performance variable was test accuracy, the test-statistic value was the $t$-value, the effect size was Cohen's $d$, and the confidence level was 95\%.

Tables~\ref{tab:MI1_ablation_ttest}-\ref{tab:MI3_ablation_ttest} present the paired $t$-test details for the ablation studies on MI1, MI2 and MI3, respectively, including sample means, standard errors, Cohen's $d$ values, $t$-values, and $p$-values. The degrees of freedom for these tests were $53$, $83$, and $71$, respectively.

\begin{table*}[htbp] \centering
	\caption{Paired $t$-test details for the ablation studies on MI1. $p$-values smaller than 0.05 are marked in bold. BN and SAM represent the two components of FedBS: local batch-specific BN and SAM optimizer.}
	\label{tab:MI1_ablation_ttest}
	\begin{tabular}{c|c|cc|cc|ccc}
		\toprule
		Model                           & FedBS vs. & Mean (FedBS)            & SE (FedBS)             & Mean (Compared) & SE (Compared) & Cohen's $d$ & $t$-value & $p$-value \\ \midrule
		\multirow{3}{*}{EEGNet}         & w/o BN,SAM    & \multirow{3}{*}{53.308} & \multirow{3}{*}{2.153} & 45.257          & 1.616         & 0.945     & 6.946   & $\boldsymbol{<0.001}$   \\
		& w/ BN  &  &  & 51.893 & 2.039 & 0.340 & 2.500 & $\boldsymbol{0.016}$ \\
		& w/ SAM &  &  & 50.134 & 1.856 & 0.496 & 3.646 & $\boldsymbol{<0.001}$ \\ \midrule
		\multirow{3}{*}{DeepConvNet}    & w/o BN,SAM   & \multirow{3}{*}{51.794} & \multirow{3}{*}{1.725} & 43.535          & 1.277         & 0.944     & 6.937   & $\boldsymbol{<0.001}$   \\
		& w/ BN  &  &  & 50.369 & 1.876 & 0.294 & 2.163 & $\boldsymbol{0.035}$ \\
		& w/ SAM &  &  & 46.276 & 1.354 & 0.830 & 6.100 & $\boldsymbol{<0.001}$ \\ \midrule
		\multirow{3}{*}{ShallowConvNet} & w/o BN,SAM    & \multirow{3}{*}{51.119} & \multirow{3}{*}{1.850} & 48.245          & 1.566         & 0.515     & 3.781   & $\boldsymbol{<0.001}$   \\
		& w/ BN  &  &  & 49.592 & 1.732 & 0.454 & 3.337 & $\boldsymbol{0.002}$ \\
		& w/ SAM &  &  & 49.733 & 1.845 & 0.375 & 2.763 & $\boldsymbol{0.008}$ \\ \bottomrule
	\end{tabular}
\end{table*}

\begin{table*}[htbp] \centering
	\caption{Paired $t$-test details for the ablation studies on MI2. $p$-values smaller than 0.05 are marked in bold. BN and SAM represent the two components of FedBS: local batch-specific BN and SAM optimizer.}
	\label{tab:MI2_ablation_ttest}
	\begin{tabular}{c|c|cc|cc|ccc}
		\toprule
		Model                           & FedBS vs. & Mean (FedBS)            & SE (FedBS)             & Mean (Compared) & SE (Compared) & Cohen's $d$ & $t$-value & $p$-value \\ \midrule
		\multirow{3}{*}{EEGNet}         & w/o BN,SAM    & \multirow{3}{*}{76.488} & \multirow{3}{*}{1.260} & 72.805          & 1.354         & 0.535     & 4.906   & $\boldsymbol{<0.001}$   \\
		& w/ BN  &  &  & 75.973 & 1.207 & 0.158 & 1.452 & $0.150$ \\
		& w/ SAM &  &  & 75.320 & 1.401 & 0.263 & 2.409 & $\boldsymbol{0.018}$ \\ \midrule
		\multirow{3}{*}{DeepConvNet}    & w/o BN,SAM    & \multirow{3}{*}{77.247} & \multirow{3}{*}{1.287} & 72.947          & 1.331         & 0.549     & 5.031   & $\boldsymbol{<0.001}$   \\
		& w/ BN  &  &  & 76.241 & 1.270 & 0.222 & 2.033 & $\boldsymbol{0.045}$ \\
		& w/ SAM &  &  & 73.883 & 1.380 & 0.506 & 4.635 & $\boldsymbol{<0.001}$ \\ \midrule
		\multirow{3}{*}{ShallowConvNet} & w/o BN,SAM    & \multirow{3}{*}{72.679} & \multirow{3}{*}{1.187} & 70.551          & 1.222         & 0.384     & 3.522   & $\boldsymbol{<0.001}$   \\
		& w/ BN  &  &  & 71.042 & 1.224 & 0.287 & 2.626 & $\boldsymbol{0.010}$ \\
		& w/ SAM &  &  & 72.432 & 1.240 & 0.111 & 1.016 & $0.313$ \\ \bottomrule
	\end{tabular}
\end{table*}

\begin{table*}[htbp] \centering
	\caption{Paired $t$-test details for the ablation studies on MI3. $p$-values smaller than 0.05 are marked in bold. BN and SAM represent the two components of FedBS: local batch-specific BN and SAM optimizer.}
	\label{tab:MI3_ablation_ttest}
	\begin{tabular}{c|c|cc|cc|ccc}
		\toprule
		Model                           & FedBS vs. & Mean (FedBS)            & SE (FedBS)             & Mean (Compared) & SE (Compared) & Cohen's $d$ & $t$-value & $p$-value \\ \midrule
		\multirow{3}{*}{EEGNet}         & w/o BN,SAM    & \multirow{3}{*}{76.375} & \multirow{3}{*}{1.484} & 72.109          & 1.206         & 0.329     & 2.793   & $\boldsymbol{0.007}$   \\
		& w/ BN  &  &  & 76.083 & 1.448 & 0.149 & 1.262  & $0.211$ \\
		& w/ SAM &  &  & 74.788 & 1.768 & 0.287 & 2.433  & $\boldsymbol{0.018}$ \\ \midrule
		\multirow{3}{*}{DeepConvNet}    & w/o BN,SAM    & \multirow{3}{*}{76.340} & \multirow{3}{*}{1.449} & 64.917          & 1.737         & 1.383     & 11.734  & $\boldsymbol{<0.001}$   \\
		& w/ BN  &  &  & 75.825 & 1.341 & 0.168 & 1.428  & $0.158$ \\
		& w/ SAM &  &  & 66.940 & 1.820 & 1.340 & 11.370 & $\boldsymbol{<0.001}$ \\ \midrule
		\multirow{3}{*}{ShallowConvNet} & w/o BN,SAM    & \multirow{3}{*}{75.680} & \multirow{3}{*}{1.420} & 74.402          & 1.454         & 0.226     & 1.914   & $0.060$   \\
		& w/ BN  &  &  & 75.163 & 1.402 & 0.166 & 1.409  & $0.163$ \\
		& w/ SAM &  &  & 75.409 & 1.426 & 0.128 & 1.089  & $0.280$ \\ \bottomrule
	\end{tabular}
\end{table*}

\end{document}